\documentclass[11pt]{article}

\usepackage{acl}

\usepackage{times}
\usepackage{latexsym}
\usepackage{tcolorbox}
\tcbuselibrary{skins}

\usepackage[T1]{fontenc}

\usepackage[utf8]{inputenc}

\usepackage{microtype}

\usepackage{inconsolata}

\usepackage{graphicx}
\usepackage{hyperref}
\usepackage{xcolor}
\usepackage{array}
\usepackage{booktabs}
\usepackage{multirow}

\newcommand{\datalink}[1]{\href{#1}{\textcolor{blue}{\underline{Link}}}}

\newcolumntype{C}[1]{>{\centering\arraybackslash}m{#1}}

\title{When Retrieval Doesn’t Help: A Large-Scale Study of Biomedical RAG}

\author{Erfan Nourbakhsh, Rocky Slavin, Ke Yang, \and Anthony Rios \\
  The University of Texas at San Antonio \\
  \texttt{\{Erfan.Nourbakhsh, Rocky.Slavin, Ke.Yang, Anthony.Rios\}@utsa.edu} \\}

\begin{document}
\maketitle
\begin{abstract}
Medical question answering is a high-stakes setting where factual errors can have serious consequences. Retrieval-augmented generation (RAG) is widely viewed as a promising solution, and prior work has reported substantial gains for large medical QA models. We revisit this assumption across a broad range of open-weight instruction-tuned models spanning 7B to 72B parameters. Across five models, ten biomedical QA datasets, four retrieval methods, and four retrieval corpora, we find that retrieval yields only small and inconsistent improvements over a no-retrieval baseline, typically within 1--2 points. In contrast, the choice of backbone model has a much larger effect than the choice of retriever or corpus, and expert and layman retrieval sources perform similarly in most settings. These results suggest that the main bottleneck is not retrieval quality alone, but the model's limited ability to use retrieved evidence effectively. Code is available here: \url{https://github.com/erfan-nourbakhsh/BioMedicalRAG}
\end{abstract}

\section{Introduction}
\label{sec:introduction}
Accurate and reliable medical question answering is a high-stakes problem, where errors can have direct consequences for patient safety. Large language models (LLMs) have recently shown strong performance on a range of biomedical question answering tasks~\cite{singhal2023large,hendrycks2020measuring,jin2021disease}. However, they remain prone to hallucination, producing fluent but factually incorrect responses~\cite{ji2023survey}, and to knowledge staleness due to their reliance on fixed training corpora. In the medical domain, these limitations are especially problematic because even small factual errors can lead to harmful downstream decisions.

Retrieval-augmented generation (RAG)~\cite{lewis2020retrieval} has become a leading approach for addressing these limitations by grounding model outputs in retrieved external evidence. By incorporating supporting documents at inference time, RAG offers a mechanism for improving factuality, transparency, and access to more current knowledge. As a result, RAG has been adopted widely in biomedical NLP, where recent work has reported substantial gains from retrieval-based methods. For example, \citet{xiong2024benchmarking} showed that MedRAG improves biomedical QA accuracy by as much as 18\% over chain-of-thought prompting, while \citet{tang2024medagents} found that multi-agent LLM systems can further improve medical reasoning performance. These findings have led to growing interest in retrieval-centered biomedical QA systems, with increasing attention to the choice of corpora, retrieval methods, and model backbones.
\begin{figure}[t]
    \centering
    \includegraphics[width=\linewidth]{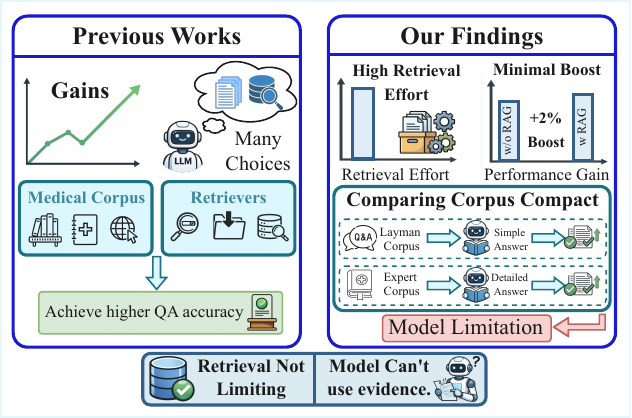}
    \caption{Overview of our motivation and main finding: across models from 7B to 72B, retrieval yields only small gains, suggesting that the main bottleneck is evidence use rather than retrieval quality.}
    \label{fig:motivation}\vspace{-1em}
\end{figure}

However, an important gap remains. Prior systematic studies of medical RAG~\cite{xiong2024benchmarking} have largely focused on large proprietary or 70B-scale models (GPT-3.5, GPT-4, Mixtral-8x7B, Llama2-70B) under zero-shot multiple-choice evaluation, leaving unclear whether their gains carry over to 7B--8B models that are far more practical under real hardware constraints. Existing evaluations have also focused primarily on expert-level biomedical questions, with little attention to consumer-health queries or community-generated retrieval sources.

In this paper, we revisit biomedical RAG under a substantially different and more comprehensive setting. We evaluate five open-weight instruction-tuned models spanning 7B to 72B parameters: Qwen2.5-7B-Instruct~\cite{yang2025qwen3}, Llama-3.1-8B-Instruct~\cite{grattafiori2024llama}, Mistral-7B-Instruct-v0.3~\cite{jiang20236g}, LLaMA-3.1-70B-Instruct~\cite{grattafiori2024llama}, and Qwen2.5-72B-Instruct~\cite{yang2025qwen3}, across ten biomedical QA datasets spanning both lay and expert questions and covering both open-ended and multiple-choice formats. We compare four retrieval methods, BM25, TF-IDF, MedCPT, and Hybrid RRF, across four retrieval corpora, including both expert biomedical resources and consumer-facing health sources: PubMed abstracts, medical textbooks, Yahoo Answers, and HealthCareMagic. We also evaluate against a no-retrieval baseline in order to isolate the contribution of retrieval itself.

Our results challenge the prevailing picture from prior studies. Across all five models, retrieval yields only small and inconsistent gains: the gap between the best retrieval configuration and the no-retrieval baseline is usually within 1--2 points (e.g., BERTScore 62.88 vs.\ 61.72 for Llama-8B, 63.23 vs.\ 61.28 for Qwen-7B), and differences across retrieval corpora are similarly modest even for the larger 70B models. By contrast, backbone model choice has a much larger effect than retriever or corpus selection, and expert versus lay retrieval sources differ by less than 2 points in most settings. Figure~\ref{fig:motivation} illustrates the key implication: the limiting factor is not retrieval quality but the generator's capacity to incorporate retrieved evidence.

Our contributions are: (1)~A large-scale evaluation of biomedical RAG covering 5 models from 7B to 72B parameters, 10 QA datasets, 4 retrieval methods, and 4 retrieval corpora. (2)~We show that retrieval yields only small and inconsistent improvements across all model scales (typically within 2 points), challenging the gains reported in prior large-model studies. (3)~We show that backbone model choice matters more than retriever or corpus choice, and provide evidence that the main bottleneck is the model's weak use of retrieved evidence.

\begin{figure*}[t]
    \centering
    \includegraphics[width=\linewidth]{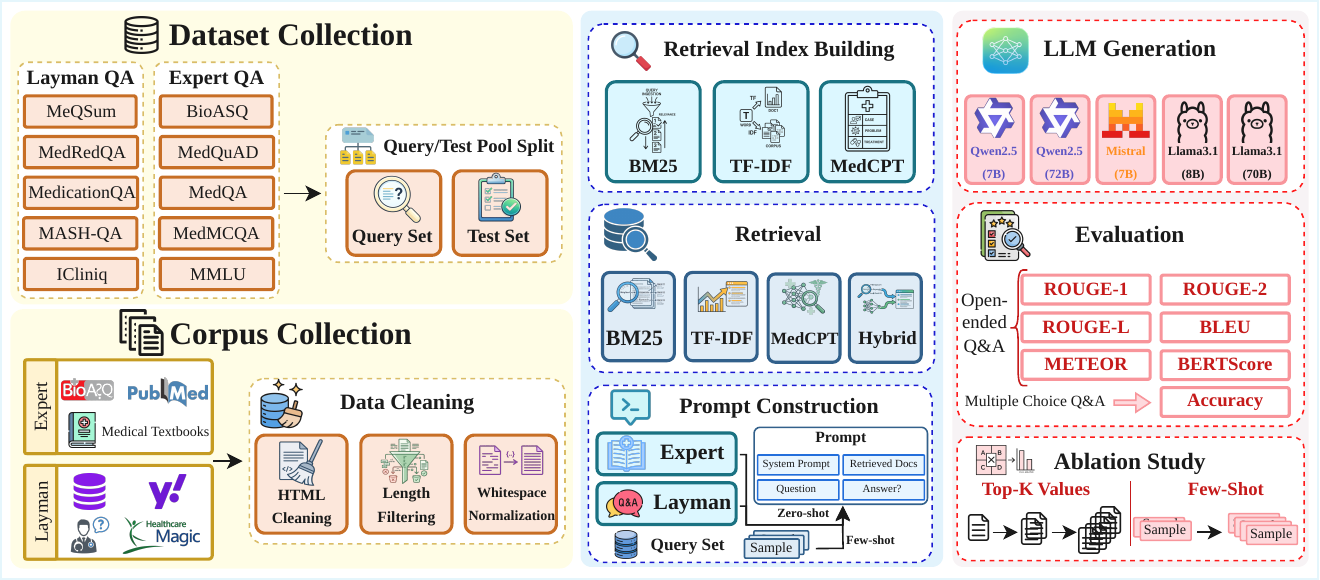}
    \caption{Experimental pipeline overview.}\vspace{-1em}
    \label{fig:pipeline}
\end{figure*}

\section{Related Work}
\label{sec:related}

\noindent \textbf{Retrieval-Augmented Generation.}
RAG was introduced by \citet{lewis2020retrieval} as a method to enhance language models on knowledge-intensive tasks by conditioning generation on documents retrieved from a non-parametric memory. The approach combines a parametric sequence-to-sequence model with a dense passage retrieval component~\cite{karpukhin2020dense} and has been extended in numerous directions, including iterative retrieval~\cite{trivedi2023interleaving,shao2023enhancing}, self-reflective retrieval~\cite{asai2023self}, and query rewriting~\cite{ma2023query}; for a broad survey of RAG paradigms and architectures, see \citet{gao2023retrieval}. In the biomedical domain, RAG has been applied to clinical decision support~\cite{xiong2024benchmarking}, scientific literature search~\cite{jin2023medcpt}, and consumer health QA~\cite{li2023chatdoctor}. However, most prior biomedical RAG studies either lack systematic comparison across retrieval configurations or are limited in dataset coverage.

\vspace{1mm} \noindent \textbf{Benchmarking Medical RAG.}
The most directly related work to ours is the MIRAGE benchmark and MedRAG toolkit by \citet{xiong2024benchmarking}, which evaluates 41 combinations of corpora, retrievers, and backbone LLMs on five medical QA datasets restricted to multiple-choice questions. MIRAGE shows that RAG can improve LLM accuracy by up to 18\% and identifies PubMed combined with BM25 or MedCPT as strong retrieval configurations. However, MIRAGE exclusively uses zero-shot prompting and evaluates primarily large models (GPT-4, GPT-3.5, Mixtral-8×7B, Llama2-70B), leaving open the question of whether these gains hold for smaller, more widely deployable models. Concurrent large-model evaluations, such as \citet{nori2023capabilities}, who find that GPT-4 surpasses the USMLE passing threshold by over 20 points even without retrieval augmentation, further underscore that model scale is a critical confound in existing medical benchmarks.
\citet{tang2024medagents} propose a zero-shot multi-agent framework achieving competitive GPT-4 performance on MMLU Medical, yet neither this nor MIRAGE examines retrieval for open-ended or consumer-health queries at the 7--8B scale. \citet{shi2023large} show that irrelevant retrieved passages can mislead LLMs, a concern especially acute for smaller models, while \citet{ovadia2024fine} find retrieval augmentation outperforms knowledge fine-tuning primarily for large models, further motivating our cross-scale evaluation.

\vspace{1mm} \noindent \textbf{Biomedical Question Answering Datasets.}
Biomedical QA has long served as a testbed for evaluating NLP systems in medicine~\cite{krithara2023bioasq,jin2021disease,pal2022medmcqa,hendrycks2020measuring}. Expert-oriented benchmarks such as BioASQ~\cite{nentidis2025overview}, MedQA-USMLE~\cite{jin2021disease}, and MedMCQA~\cite{pal2022medmcqa} test clinical and examination-level knowledge, while consumer-health datasets such as MeQSum~\cite{abacha2019summarization}, MedRedQA~\cite{nguyen2023medredqa}, MedicationQA~\cite{abacha2019bridging}, MASH-QA~\cite{zhu2020question}, and ChatDoctor-iCliniq~\cite{li2023chatdoctor} reflect more informal, everyday health information needs. MedQuAD~\cite{ben2019question} and MMLU Medical~\cite{hendrycks2020measuring} bridge the two groups by covering structured NIH-sourced QA and standardised medical knowledge. Despite this rich landscape, most RAG studies focus on MCQ-format expert benchmarks and omit the open-ended and layman query types that constitute the bulk of real-world health information needs, a gap we directly address.

\vspace{1mm} \noindent \textbf{Retrieval Methods in Biomedical NLP.}
Sparse retrieval methods have been dominant in biomedical information retrieval. BM25~\cite{robertson2009probabilistic}, a probabilistic bag-of-words ranking function, remains a strong baseline and is adopted as the primary retriever in MedRAG~\cite{xiong2024benchmarking}. TF-IDF~\cite{sparck1972statistical}, a simpler precursor that models term specificity without BM25's saturation and length normalisation, provides a useful lower bound for sparse retrieval.

Dense retrieval with domain-adapted encoders has gained traction. MedCPT~\cite{jin2023medcpt} was trained contrastively on large-scale PubMed search logs and demonstrates strong zero-shot biomedical retrieval, outperforming general-domain encoders on medical tasks. Fusion methods such as Reciprocal Rank Fusion (RRF)~\cite{cormack2009reciprocal} combine sparse and dense ranked lists and have been shown to outperform individual retrievers without requiring additional training. While MedRAG includes RRF as a configuration, it does not systematically isolate the contribution of each component retriever across diverse query types and corpora, which we do in this study.

\section{Experiments}
\label{sec:experiments}
Our experiments systematically compare sparse, dense, and hybrid retrieval strategies across four corpora, ten QA datasets spanning expert and layman health queries, and five open-weight instruction-tuned models ranging from 7B to 72B parameters. Figure~\ref{fig:pipeline} provides a visual overview of the full experimental pipeline.

\vspace{1mm} \noindent \textbf{Evaluation Dataset and Knowledge Base.}
\label{sec:datasets-main}

\vspace{1mm} \noindent \textit{\textbf{Evaluation Datasets.}}
\label{sec:eval-datasets}
We evaluate across ten biomedical and consumer-health question answering datasets grouped into two user types: \textit{layman} datasets reflecting everyday consumer-health language, and \textit{expert} datasets targeting biomedical professionals or medical students. Dataset statistics and split sizes are summarised in Table~\ref{tab:query-datasets} in Appendix~\ref{sec:appendix-datasets}. For all datasets, examples lacking a question or a reference answer are discarded before any split is finalised, and whenever random sampling is needed it is performed with a fixed seed of 42.

\noindent \textit{\textbf{Layman datasets.}}
\textbf{MeQSum}~\cite{abacha2019summarization} contains 1,000 consumer health questions from the U.S.\ National Library of Medicine.
Following \citet{zhang2022focus}, we reserve 500 examples for evaluation and use the remaining 500 as the few-shot query pool.
\textbf{MedRedQA}~\cite{nguyen2023medredqa} provides over 51,000 consumer question--physician answer pairs from Reddit's \texttt{/r/AskDocs}; we sample 1,000 evaluation examples from the official test split (5,099 examples) and combine the training (40,792) and validation (5,100) splits into the query pool.
\textbf{MedicationQA}~\cite{abacha2019bridging} contains 690 real consumer medication questions; we randomly sample 500 for evaluation and retain the remaining 189 as the query pool.
\textbf{MASH-QA}~\cite{zhu2020question} offers over 34,000 WebMD-derived healthcare Q\&A pairs; we randomly sample 1,000 examples from the official test file (2,614 entries) and use the full training set (19,989 examples) as the query pool.
\textbf{ChatDoctor-iCliniq}~\cite{li2023chatdoctor} comprises 7,321 real patient--physician conversations from iCliniq.com; we randomly sample 1,000 for evaluation and retain the remaining 6,321 as the query pool.

\vspace{1mm} \noindent \textit{\textbf{Expert datasets.}}
\textbf{BioASQ Task~B}~\cite{nentidis2025overview} provides expert biomedical questions grounded in PubMed literature; following the official benchmark protocol, we use the Task~13B golden test set (restricted to summary-type questions, 80 examples) and the Task~13B training set (1,283 examples) as the query pool.
\textbf{MedQuAD}~\cite{ben2019question} contains 47,457 medical Q\&A pairs from 12 NIH websites, of which 16,407 are publicly available; we randomly sample 1,000 for evaluation and use the remaining 15,407 as the query pool.
\textbf{MedQA-USMLE}~\cite{jin2021disease} provides USMLE clinical vignette MCQs; we use the official test split (1,273 examples) for evaluation and the official training split (10,178 examples) as the query pool.
\textbf{MedMCQA}~\cite{pal2022medmcqa} contains 194k+ MCQs from AIIMS and NEET PG medical entrance exams; as the official test split is unlabelled, we randomly sample 1,000 from the validation set (6,150 examples) for evaluation and use the full training set (182,822 examples) as the query pool.
\textbf{MMLU Medical}~\cite{hendrycks2020measuring}: following \citet{tang2024medagents}, we restrict to six medical sub-tasks, \textit{anatomy}, \textit{clinical\_knowledge}, \textit{college\_biology}, \textit{college\_medicine}, \textit{medical\_genetics}, and \textit{professional\_medicine}, totalling 1,242 examples. Roughly 100 examples per sub-task (600 in total) are used for evaluation; the remaining 642 form the query pool.

\vspace{1mm} \noindent \textbf{Knowledge Bases.}
\label{sec:corpora}
We build four retrieval corpora covering both expert biomedical and layman health domains, as summarised in Table~\ref{tab:retrieval-corpora} in Appendix~\ref{sec:appendix-datasets}. All corpora are indexed as whole records without further chunking. For Q\&A-style corpora (Yahoo Answers and HealthCareMagic), each document concatenates the question or title with the corresponding answer body.

\vspace{1mm} \noindent \textit{\textbf{BioASQ / PubMed~\cite{krithara2023bioasq}}} consists of 16.2 million PubMed abstracts with human-assigned MeSH annotations and serves as the primary expert-domain knowledge base.

\vspace{1mm} \noindent \textit{\textbf{Medical Textbooks~\cite{xiong2024benchmarking}}} provides 125,847 retrieval-friendly chunks ($\leq$1,000 characters each) drawn from 18 authoritative biomedical textbooks spanning anatomy, physiology, pharmacology, pathology, and clinical medicine.

\vspace{1mm} \noindent \textit{\textbf{Yahoo Answers~\cite{yahoo_webscope_2009}}} is an open-domain community Q\&A corpus; from the original 1.4 million records we retain 1,238,506 after quality filtering, discarding entries whose answer body contains fewer than five words or whose combined question--answer text falls below ten words.

\vspace{1mm} \noindent \textit{\textbf{HealthCareMagic~\cite{li2023chatdoctor}}} contains 112,165 real-world patient symptom queries paired with detailed physician responses across more than ten clinical specialties.

\begin{table*}[t]
\centering
\resizebox{\textwidth}{!}{%
\begin{tabular}{llcccccccc}
\toprule
\textbf{Model} & \textbf{Retrieval Dataset} & \textbf{BioASQ} & \textbf{ChatDoctor/iCliniq} & \textbf{MashQA} & \textbf{MedicationQA} & \textbf{MedQuAD} & \textbf{MedRedQA} & \textbf{MeQSum} & \textbf{Average} \\
\midrule
\multirow{5}{*}{LLaMA3.1-8B}
 & w/o RAG            & 21.65 & 12.15 & 14.01 & 12.29 & \textbf{16.75} & 8.76 & 5.82 & 13.06 \\\cmidrule(lr){3-10}
 & BioASQ             & \textbf{27.43} & 12.54 & \textbf{15.26} & \textbf{12.43} & 16.11 & \textbf{8.84} & \textbf{7.04} & \textbf{14.24} \\
 & HealthCareMagic    & 19.90 & \textbf{12.71} & 14.41 & 11.87 & 16.22 & 8.76 & 6.43 & 12.90 \\
 & Medical Textbooks  & 22.07 & 12.54 & 15.04 & 12.30 & 15.47 & \textbf{8.84} & 6.30 & 13.22 \\
 & Yahoo Answers      & 20.05 & 12.48 & 15.04 & 11.47 & 15.87 & 8.72 & 6.03 & 12.81 \\
\midrule
\multirow{5}{*}{LLaMA3.1-70B}
 & w/o RAG            & 21.93 & 13.18 & \textbf{17.19} & 14.61 & \textbf{18.35} & 9.07 & 5.23 & 14.22 \\\cmidrule(lr){3-10}
 & BioASQ             & \textbf{28.98} & 13.10 & 15.95 & 14.03 & 15.71 & 9.13 & 5.73 & \textbf{14.66} \\
 & HealthCareMagic    & 23.58 & 13.35 & 16.28 & \textbf{15.27} & 17.63 & \textbf{9.29} & 5.70 & 14.44 \\
 & Medical Textbooks  & 24.71 & 13.14 & 15.73 & 14.96 & 15.50 & 9.21 & 5.72 & 14.14 \\
 & Yahoo Answers      & 23.68 & \textbf{13.30} & 16.18 & 14.78 & 16.22 & 9.26 & \textbf{5.88} & 14.19 \\
\midrule
\multirow{5}{*}{Mistral-7B}
 & w/o RAG            & 22.55 & 13.27 & 15.11 & 12.62 & 17.05 & 9.05 & 5.82 & 13.64 \\\cmidrule(lr){3-10}
 & BioASQ             & \textbf{26.55} & 13.49 & 15.50 & 12.78 & 16.81 & 9.39 & 6.57 & \textbf{14.44} \\
 & HealthCareMagic    & 24.09 & \textbf{13.91} & 15.93 & 12.78 & \textbf{17.24} & 9.46 & 6.42 & 14.26 \\
 & Medical Textbooks  & 22.81 & 13.45 & 15.30 & \textbf{12.90} & 16.45 & 9.28 & 6.41 & 13.80 \\
 & Yahoo Answers      & 23.96 & 13.68 & \textbf{16.05} & 13.15 & 17.21 & \textbf{9.56} & \textbf{6.66} & 14.32 \\
\midrule
\multirow{5}{*}{Qwen2.5-7B}
 & w/o RAG            & 20.95 & 12.36 & 15.13 & 12.24 & 15.80 & 8.47 & 5.41 & 12.91 \\\cmidrule(lr){3-10}
 & BioASQ             & \textbf{23.93} & 12.46 & 15.59 & 12.41 & \textbf{16.29} & 8.54 & 5.69 & \textbf{13.56} \\
 & HealthCareMagic    & 21.21 & 12.42 & 15.20 & 11.89 & 16.07 & \textbf{8.58} & 5.61 & 13.00 \\
 & Medical Textbooks  & 21.99 & \textbf{12.49} & 15.59 & \textbf{12.42} & 15.83 & 8.56 & \textbf{5.87} & 13.25 \\
 & Yahoo Answers      & 21.29 & 12.43 & \textbf{15.74} & 12.30 & 16.12 & 8.63 & 5.86 & 13.20 \\
\midrule
\multirow{5}{*}{Qwen2.5-72B}
 & w/o RAG            & 22.22 & 12.86 & 16.03 & 12.23 & 17.17 & 8.92 & 5.47 & 13.56 \\\cmidrule(lr){3-10}
 & BioASQ             & \textbf{24.52} & 12.83 & 15.83 & 12.21 & 17.38 & 8.97 & 5.61 & \textbf{13.91} \\
 & HealthCareMagic    & 23.60 & 12.90 & 15.68 & 11.76 & 17.67 & \textbf{9.02} & 5.43 & 13.72 \\
 & Medical Textbooks  & 23.69 & \textbf{12.91} & 15.92 & \textbf{12.60} & 17.32 & 8.98 & 5.61 & 13.86 \\
 & Yahoo Answers      & 23.60 & 12.89 & \textbf{16.18} & 12.10 & \textbf{17.80} & \textbf{9.02} & \textbf{5.71} & 13.90 \\
\bottomrule
\end{tabular}%
}
\caption{ROUGE-L by model and retrieval corpus (open-ended datasets).}\vspace{-1em}
\label{tab:results_rougel}
\end{table*}

\vspace{1mm} \noindent \textbf{Retrieval Approaches.}
\label{sec:retrieval}
We compare four retrieval strategies that differ in their document and query representations.

\vspace{1mm} \noindent \textit{\textbf{BM25~\cite{robertson2009probabilistic}}} is a classic sparse probabilistic retrieval model that scores documents by the weighted overlap of query terms, applying a term-frequency saturation function and a document-length normalisation penalty. BM25 has long served as a strong baseline for ad-hoc retrieval and remains competitive with many neural approaches.
We adopt BM25 parameters $k_1{=}0.9$ and $b{=}0.4$, and apply a title-boost factor of 2 by repeating title tokens at indexing time to approximate field-weighted BM25F scoring.

\vspace{1mm} \noindent \textit{\textbf{TF-IDF~\cite{sparck1972statistical}}} represents both documents and queries as bag-of-words vectors weighted by term frequency--inverse document frequency, and ranks candidates by cosine similarity.
Unlike BM25, TF-IDF applies no term-frequency saturation or document-length penalty, making it a simpler baseline for sparse lexical matching.
We build a TF-IDF index with a vocabulary capped at 50,000 features and standard English stop-word removal.

\vspace{1mm} \noindent \textit{\textbf{MedCPT~\cite{jin2023medcpt}}} is a biomedical dense retrieval model consisting of a \textit{query encoder} and an \textit{article encoder} trained contrastively on large-scale PubMed user search logs.
Documents are encoded offline by the article encoder and stored as L2-normalised embeddings; at query time, the query encoder produces a query embedding and retrieval proceeds by maximum inner-product search.
By capturing semantic similarity beyond exact term overlap, MedCPT is particularly well-suited to the biomedical domain.

\vspace{1mm} \noindent \textit{\textbf{Hybrid BM25 + MedCPT via RRF~\cite{cormack2009reciprocal}}}
combines the BM25 and MedCPT ranked lists using Reciprocal Rank Fusion (RRF).
Each document $d$ ranked at position $r$ in a ranked list receives a score $\frac{1}{k{+}r}$; with $k{=}60$, and the scores are summed across both lists.
The final ranking is by descending combined RRF score.
RRF is parameter-light and has been shown to consistently outperform individual rankers as well as more complex score-fusion methods~\cite{cormack2009reciprocal}.

For all retrieval conditions, we retrieve the top $k{=}5$ documents and concatenate them as the retrieved context prepended to the generator prompt.

\vspace{1mm} \noindent \textbf{Implementation Details.}
\label{sec:implementation}
All generation experiments are conducted with five open-source instruction-tuned models spanning two scales. The 7--8B models are \textbf{Qwen2.5-7B-Instruct}~\cite{yang2025qwen3}, \textbf{Llama-3.1-8B-Instruct}~\cite{grattafiori2024llama}, and \textbf{Mistral-7B-Instruct-v0.3}~\cite{jiang20236g}, each servable on a single GPU. The 70B-scale models are \textbf{LLaMA-3.1-70B-Instruct}~\cite{grattafiori2024llama} and \textbf{Qwen2.5-72B-Instruct}~\cite{yang2025qwen3}, which serve as large-scale reference points to contextualise the small-model results.

All models are run in half-precision (FP16) with greedy decoding and a maximum of 300 newly generated tokens per response.

\vspace{1mm} \noindent \textbf{Experimental Setting.}
\label{sec:exp-setting}
Each experimental condition is defined by a triple (\textit{retriever}, \textit{corpus}, \textit{query dataset}). The retriever dimension covers five options: \textit{No retrieval} (baseline), BM25, TF-IDF, MedCPT, and Hybrid (BM25 + MedCPT via RRF). The corpus dimension covers four knowledge bases: BioASQ/PubMed and Medical Textbooks as expert corpora, and Yahoo Answers and HealthCareMagic as layman corpora (for the baseline condition both retriever and corpus are set to none). The query dimension covers the ten datasets described in Section~\ref{sec:eval-datasets}, split evenly between layman (MeQSum, MedRedQA, MedicationQA, MASH-QA, ChatDoctor-iCliniq) and expert (BioASQ Task~B, MedQuAD, MedQA-USMLE, MedMCQA, MMLU Medical) user types.

For the \textit{w/o RAG} condition the model receives only the question in its prompt, with no retrieved context. For retrieval-augmented conditions, the top-$k$ retrieved passages are prepended to the question in a fixed prompt template. Each condition is run independently for every generator model, and all per-dataset query pools described in Section~\ref{sec:eval-datasets} are also available for few-shot prompting ablations. The complete set of conditions spans every combination of retriever, corpus, and query dataset, yielding a large-scale cross-model, cross-retriever, cross-dataset evaluation.

\section{Results}
\label{sec:results}

We present results separately for open-ended QA, evaluated with ROUGE-L as the primary metric (ROUGE-1, ROUGE-2, METEOR, BLEU, and BERTScore in Appendix~\ref{sec:appendix-full-results}), and for multiple-choice QA, evaluated with accuracy.

\vspace{1mm} \noindent \textbf{Open-ended QA.}
\label{sec:results-openended}
Table~\ref{tab:results_rougel} reports ROUGE-L across seven open-ended datasets (five layman and two expert), averaged over all retrieval conditions per corpus. Across all models, retrieval yields small and inconsistent improvements over the no-retrieval baseline. The largest gains appear on the BioASQ open-ended task, where the BioASQ/PubMed corpus consistently provides the strongest lift: for example, LLaMA-3.1-8B improves from 21.65 to 27.43 ROUGE-L. However, for the remaining six datasets, changes from the baseline are typically under 1 ROUGE-L point and often negative. Averaged across all seven datasets, the maximum retrieval benefit over no-retrieval is 1.18 points (LLaMA-3.1-8B: 13.06 baseline vs.\ 14.24 with BioASQ); for all other models the gain is smaller still. ROUGE-1, ROUGE-2, METEOR, BLEU, and BERTScore results (Appendix~\ref{sec:appendix-full-results}) show the same pattern.

A consistent observation is that backbone model choice matters far more than retrieval configuration. Mistral-7B lags behind both LLaMA-3.1-8B and Qwen2.5-7B regardless of the retrieval setup, and the 70B-scale models (LLaMA-3.1-70B, Qwen2.5-72B) are consistently stronger than all 7--8B variants. The gap between any two retrieval conditions for the same model is almost always smaller than the gap between two different backbone models using the same conditions. Expert and layman retrieval corpora produce similar results in most open-ended settings, differing by less than 1 ROUGE-L point on average.

\begin{table}[t]
\centering
\footnotesize
\setlength{\tabcolsep}{3pt}
\renewcommand{\arraystretch}{0.95}
\begin{tabular}{l l c c c c}
\toprule
\textbf{Model} & \textbf{Data} & \textbf{MCQA} & \textbf{MQA} & \textbf{MMLU} & \textbf{Avg} \\
\midrule
\multirow{5}{*}{LLaMA3.1-8B}
 & w/o RAG  & \textbf{80.8} & 83.8 & \textbf{83.7} & \textbf{82.8} \\
 & BioASQ   & 75.9 & \textbf{84.6} & 82.3 & 80.9 \\
 & HCM      & 73.7 & 84.2 & 74.0 & 77.3 \\
 & Textbook & 74.1 & 84.1 & 83.3 & 80.5 \\
 & Yahoo    & 74.6 & 83.7 & 81.5 & 79.9 \\
\midrule
\multirow{5}{*}{LLaMA3.1-70B}
 & w/o RAG  & \textbf{81.0} & 89.1 & 89.2 & 86.4 \\
 & BioASQ   & 79.8 & 90.8 & \textbf{90.2} & \textbf{86.9} \\
 & HCM      & 78.9 & 80.5 & 87.5 & 82.3 \\
 & Textbook & 80.2 & 81.1 & 89.5 & 83.6 \\
 & Yahoo    & 79.1 & \textbf{91.7} & 88.7 & 86.5 \\
\midrule
\multirow{5}{*}{Mistral-7B}
 & w/o RAG  & \textbf{72.6} & \textbf{77.9} & \textbf{76.7} & \textbf{75.7} \\
 & BioASQ   & 61.3 & 72.4 & 72.2 & 68.6 \\
 & HCM      & 63.3 & 73.2 & 72.0 & 69.5 \\
 & Textbook & 66.0 & 73.7 & 77.3 & 72.3 \\
 & Yahoo    & 65.5 & 74.0 & 74.2 & 71.2 \\
\midrule
\multirow{5}{*}{Qwen2.5-7B}
 & w/o RAG  & \textbf{80.0} & \textbf{83.7} & \textbf{86.3} & \textbf{83.3} \\
 & BioASQ   & 75.8 & 81.1 & 82.2 & 79.7 \\
 & HCM      & 74.9 & 81.4 & 82.8 & 79.7 \\
 & Textbook & 76.6 & 81.1 & 85.7 & 81.1 \\
 & Yahoo    & 77.5 & 81.4 & 85.2 & 81.4 \\
\midrule
\multirow{5}{*}{Qwen2.5-72B}
 & w/o RAG  & \textbf{82.5} & 81.9 & \textbf{92.5} & \textbf{85.6} \\
 & BioASQ   & 80.1 & 81.6 & 91.2 & 84.3 \\
 & HCM      & 77.3 & \textbf{85.8} & 90.7 & 84.6 \\
 & Textbook & 80.1 & 83.5 & 91.2 & 84.9 \\
 & Yahoo    & 79.8 & 81.8 & 90.5 & 84.0 \\
\bottomrule
\end{tabular}
\caption{Accuracy by model and retrieval corpus. MCQA denotes MedMCQA, MQA denotes MedQA, and HCM denotes HealthCareMagic.}\vspace{-1em}
\label{tab:results_accuracy}
\end{table}

\vspace{1mm} \noindent \textbf{Multiple-choice QA.}
\label{sec:results-mcq}
Table~\ref{tab:results_accuracy} reports accuracy grouped by dataset subset (MCQA: MedQA + MedMCQA; QA: open-ended expert; MMLU: six MMLU medical subjects). For smaller models (LLaMA-3.1-8B, Mistral-7B, Qwen2.5-7B), retrieval frequently \emph{hurts} accuracy relative to the no-retrieval baseline. Mistral-7B drops from 75.7 to 68.6--72.3 across all retrieval corpora. The larger models (LLaMA-3.1-70B, Qwen2.5-72B) are more robust, maintaining accuracy within 1--2 points of the baseline across all conditions, but still show no consistent gain. As with the open-ended setting, backbone choice dominates: Qwen2.5-72B's no-retrieval accuracy of 85.6 exceeds the best retrieval configuration of any 7B model by over 2 points.

\vspace{1mm} \noindent \textbf{Effect of Retrieval Method.}
\label{sec:results-retriever}
Table~\ref{tab:rougel_retrieval} reports accuracy averaged across the three close-ended datasets from Table~\ref{tab:results_accuracy} (MedMCQA, MedQA-USMLE, and MMLU Medical), broken down by retrieval method rather than corpus. Table~\ref{tab:rougel_retrieval_open} reports ROUGE-L averaged across the seven open-ended datasets from Table~\ref{tab:results_rougel} (BioASQ, ChatDoctor-iCliniq, MashQA, MedicationQA, MedQuAD, MedRedQA, and MeQSum), again broken down by retrieval method. Together, these two tables allow direct comparison of BM25, Hybrid (RRF), MedCPT, and TF-IDF across question types and retrieval corpora. Differences among methods are within 1--2 points for any model--corpus combination. The Hybrid retriever shows marginal advantages in several configurations, but no method consistently dominates. MedCPT, despite domain-specific training, does not systematically outperform lexical BM25. Full per-metric breakdowns by retriever type (ROUGE-1, ROUGE-2, METEOR, BLEU, BERTScore) are in Appendix~\ref{sec:appendix-full-results}.

\begin{table}[t]
\centering
\resizebox{\columnwidth}{!}{%
\begin{tabular}{llcccc}
\toprule
\textbf{Model} & \textbf{Retrieval Dataset} & \textbf{BM25} & \textbf{Hybrid} & \textbf{MedCPT} & \textbf{TF-IDF} \\
\midrule
\multirow{5}{*}{LLaMA3.1-8B}
 & BioASQ            & 79.34 & 79.57 & \textbf{80.94} & 80.65 \\
 & HealthCareMagic   & 72.62 & 74.59 & 74.88 & \textbf{76.97} \\
 & Medical Textbooks & 79.47 & \textbf{80.34} & 79.53 & 79.08 \\
 & Yahoo Answers     & 79.27 & \textbf{79.74} & 77.91 & 77.14 \\
\cmidrule(lr){2-6}
 & Average           & 77.68 & \textbf{78.56} & 78.32 & 78.46 \\
\midrule
\multirow{5}{*}{LLaMA3.1-70B}
 & BioASQ            & 85.56 & 82.11 & \textbf{86.52} & 85.06 \\
 & HealthCareMagic   & 81.92 & 80.99 & 81.52 & \textbf{82.14} \\
 & Medical Textbooks & 81.54 & 82.87 & 82.68 & \textbf{83.29} \\
 & Yahoo Answers     & 85.51 & 81.69 & 81.32 & \textbf{85.77} \\
\cmidrule(lr){2-6}
 & Average           & 83.64 & 81.91 & 83.01 & \textbf{84.06} \\
\midrule
\multirow{5}{*}{Mistral-7B}
 & BioASQ            & 66.82 & 68.13 & \textbf{68.52} & 67.73 \\
 & HealthCareMagic   & 68.36 & 68.81 & \textbf{69.12} & 68.64 \\
 & Medical Textbooks & 70.66 & \textbf{71.47} & 70.52 & 69.37 \\
 & Yahoo Answers     & 68.59 & 69.43 & 69.72 & \textbf{70.94} \\
\cmidrule(lr){2-6}
 & Average           & 68.61 & 69.46 & \textbf{69.47} & 69.17 \\
\midrule
\multirow{5}{*}{Qwen2.5-7B}
 & BioASQ            & 78.37 & 78.88 & \textbf{79.61} & 79.49 \\
 & HealthCareMagic   & 78.39 & 78.55 & 77.87 & \textbf{78.84} \\
 & Medical Textbooks & \textbf{80.28} & 80.15 & \textbf{80.28} & 79.39 \\
 & Yahoo Answers     & 80.52 & 80.03 & \textbf{80.81} & 79.69 \\
\cmidrule(lr){2-6}
 & Average           & 79.39 & 79.40 & \textbf{79.64} & 79.35 \\
 \midrule
 \multirow{5}{*}{Qwen2.5-72B}
 & BioASQ            & 83.05 & \textbf{83.86} & 83.58 & 83.16 \\
 & HealthCareMagic   & 82.34 & 82.70 & 82.80 & \textbf{83.84} \\
 & Medical Textbooks & \textbf{84.72} & 84.37 & 83.04 & 82.63 \\
 & Yahoo Answers     & 83.22 & 83.44 & \textbf{83.69} & 82.90 \\
\cmidrule(lr){2-6}
 & Average           & 83.33 & \textbf{83.59} & 83.28 & 83.13 \\
\bottomrule
\end{tabular}%
}
\caption{Accuracy by retrieval method (close-ended datasets).}
\label{tab:rougel_retrieval}
\end{table}

\section{Ablation Study}
\label{sec:ablation}

We conduct two ablations to understand how retrieval depth and few-shot context affect performance. Both use a stratified subset of the test queries with BM25 retrieval from BioASQ. Additional open-ended metric trends are visualised in Appendix~\ref{sec:appendix-ablations} (Figures~\ref{fig:open-few-shot-appeindex} and~\ref{fig:open-k-value-appeindex}).

\begin{table}[t]
\centering
\resizebox{\columnwidth}{!}{%
\begin{tabular}{llccccc}
\toprule
\textbf{Model} & \textbf{Retrieval Dataset} & \textbf{BM25} & \textbf{Hybrid} & \textbf{MedCPT} & \textbf{TF-IDF} \\
\midrule
\multirow{5}{*}{LLaMA3.1-8B}
 & BioASQ            & 13.93 & \textbf{14.10} & 13.57 & 13.17 \\
 & HealthCareMagic   & 12.47 & \textbf{12.80} & 12.58 & 12.71 \\
 & Medical Textbooks & 12.90 & 13.01 & \textbf{13.08} & 12.93 \\
 & Yahoo Answers     & 12.59 & \textbf{12.73} & 12.58 & 12.23 \\
\cmidrule(lr){2-6}
 & Average           & 12.98 & \textbf{13.16} & 12.95 & 12.76 \\
\midrule
\multirow{5}{*}{LLaMA3.1-70B}
 & BioASQ            & 13.95 & \textbf{14.58} & 13.78 & 13.59 \\
 & HealthCareMagic   & 14.19 & 14.31 & \textbf{14.39} & 14.21 \\
 & Medical Textbooks & 13.57 & 13.83 & \textbf{14.00} & 13.85 \\
 & Yahoo Answers     & 13.68 & \textbf{13.97} & \textbf{13.97} & 13.41 \\
\cmidrule(lr){2-6}
 & Average           & 13.85 & \textbf{14.17} & 14.04 & 13.77 \\
\midrule
\multirow{5}{*}{Mistral-7B}
 & BioASQ            & 14.19 & \textbf{14.34} & 13.91 & 13.92 \\
 & HealthCareMagic   & 14.08 & 14.11 & 14.01 & \textbf{14.14} \\
 & Medical Textbooks & 13.71 & \textbf{13.74} & 13.59 & 13.67 \\
 & Yahoo Answers     & 14.14 & 14.16 & 14.13 & \textbf{14.20} \\
\cmidrule(lr){2-6}
 & Average           & 14.03 & \textbf{14.09} & 13.91 & 13.98 \\
\midrule
\multirow{5}{*}{Qwen2.5-7B}
 & BioASQ            & 13.26 & \textbf{13.49} & 13.23 & 12.97 \\
 & HealthCareMagic   & 12.80 & \textbf{12.90} & 12.87 & \textbf{12.90} \\
 & Medical Textbooks & 12.98 & \textbf{13.22} & 13.04 & 13.06 \\
 & Yahoo Answers     & 12.99 & \textbf{13.13} & \textbf{13.13} & 13.05 \\
\cmidrule(lr){2-6}
 & Average           & 13.01 & \textbf{13.19} & 13.07 & 13.00 \\
 \midrule
 \multirow{5}{*}{Qwen2.5-72B}
 & BioASQ            & 13.80 & \textbf{13.82} & 13.73 & 13.75 \\
 & HealthCareMagic   & 13.57 & 13.56 & 13.65 & \textbf{13.70} \\
 & Medical Textbooks & \textbf{13.81} & 13.74 & 13.61 & 13.77 \\
 & Yahoo Answers     & 13.63 & 13.72 & 13.84 & \textbf{13.85} \\
\cmidrule(lr){2-6}
 & Average           & 13.70 & 13.71 & 13.71 & \textbf{13.77} \\
\bottomrule
\end{tabular}%
}
\caption{ROUGE-L by retrieval method (open-ended datasets).}
\label{tab:rougel_retrieval_open}
\end{table}

\begin{figure}[t]
    \centering
    \includegraphics[width=\linewidth]{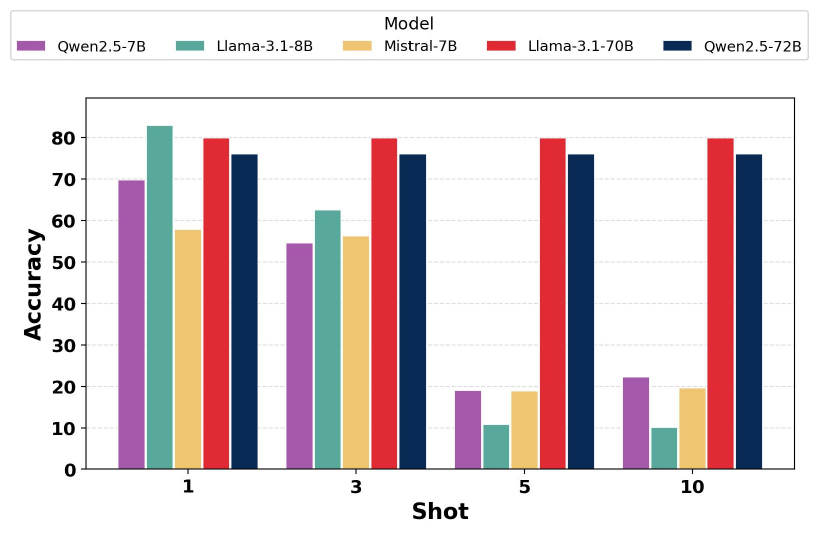}
    \caption{Close-ended accuracy across shot counts (1, 3, 5, 10).}
    \label{fig:close-few-shot}
\end{figure}

\begin{figure}[t]
    \centering
    \includegraphics[width=\linewidth]{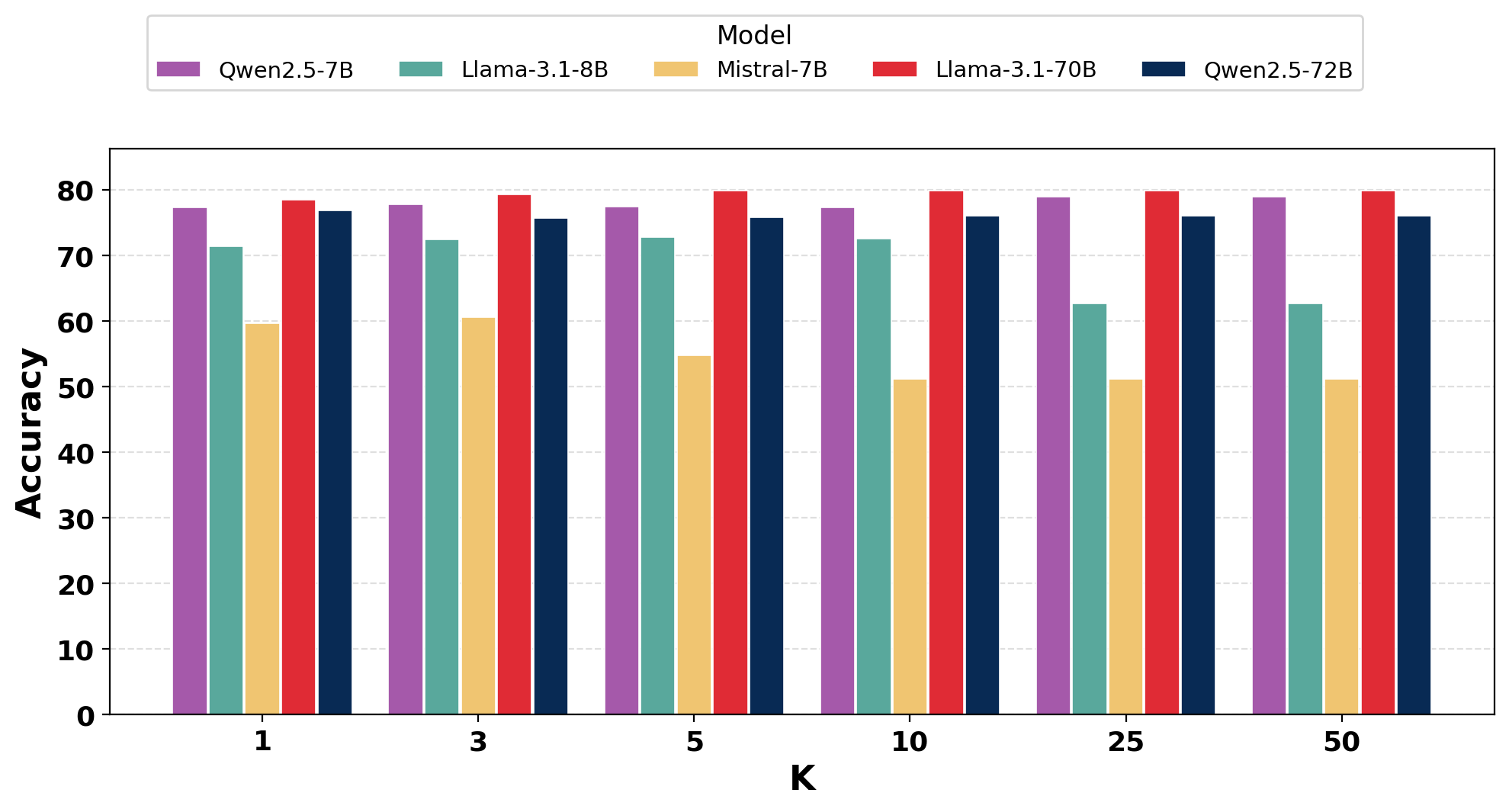}
    \caption{Close-ended accuracy across top-$k$ (1, 3, 5, 10, 25, 50).}
    \label{fig:close-k-value}
\end{figure}

\begin{figure}[t]
    \centering
    \includegraphics[width=\linewidth]{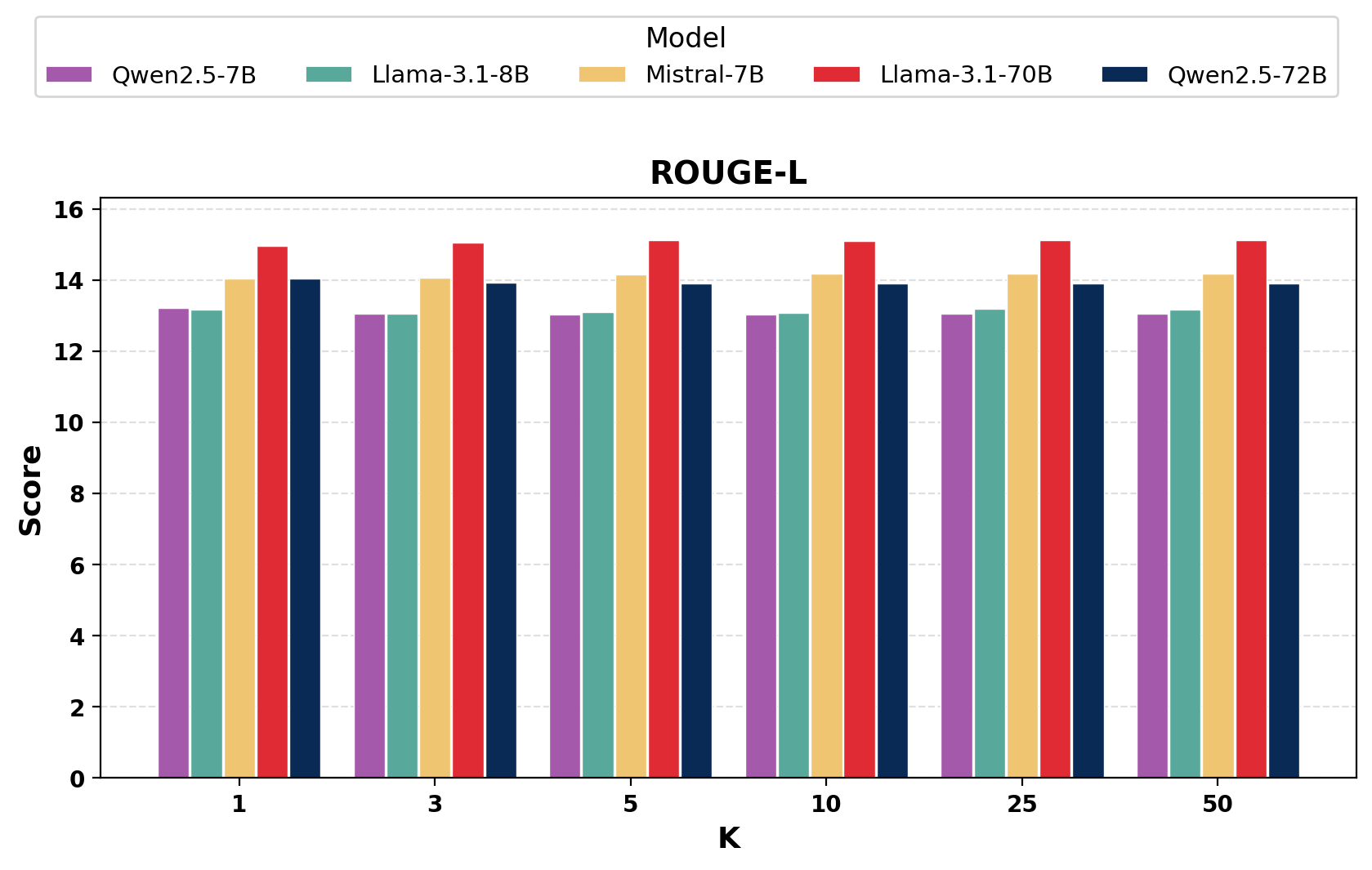}
    \caption{Open-ended ROUGE-L across top-$k$ values.}\vspace{-1em}
    \label{fig:open-k-value-main}
\end{figure}

\vspace{1mm} \noindent \textbf{Number of Retrieved Documents (Top-$k$).}
\label{sec:ablation-k}
Figures~\ref{fig:close-k-value} and~\ref{fig:open-k-value-main} show accuracy and ROUGE-L as $k$ varies over $\{1, 3, 5, 10, 25, 50\}$. For open-ended metrics, performance reaches a plateau by $k{=}5$: ROUGE-L changes by less than 0.2 points between $k{=}5$ and $k{=}50$ for all models, indicating that additional retrieved documents add no useful signal once the context budget is satisfied. For close-ended accuracy the picture is less uniform: LLaMA-3.1-8B peaks at $k{=}5$ (72.83\%) before declining, while Qwen2.5-7B and LLaMA-3.1-70B reach their best performance at $k{\geq}25$. Mistral-7B declines steadily after $k{=}3$, reaching 51.22\% at $k{\geq}25$. These results confirm that $k{=}5$ is a reasonable default: it matches or closely approaches the optimum for most models while keeping context length manageable. Additional open-ended metric trends across all $k$ values are shown in Figure~\ref{fig:open-k-value-appeindex} in the appendix.

\begin{figure}[ht]
    \centering
    \includegraphics[width=\linewidth]{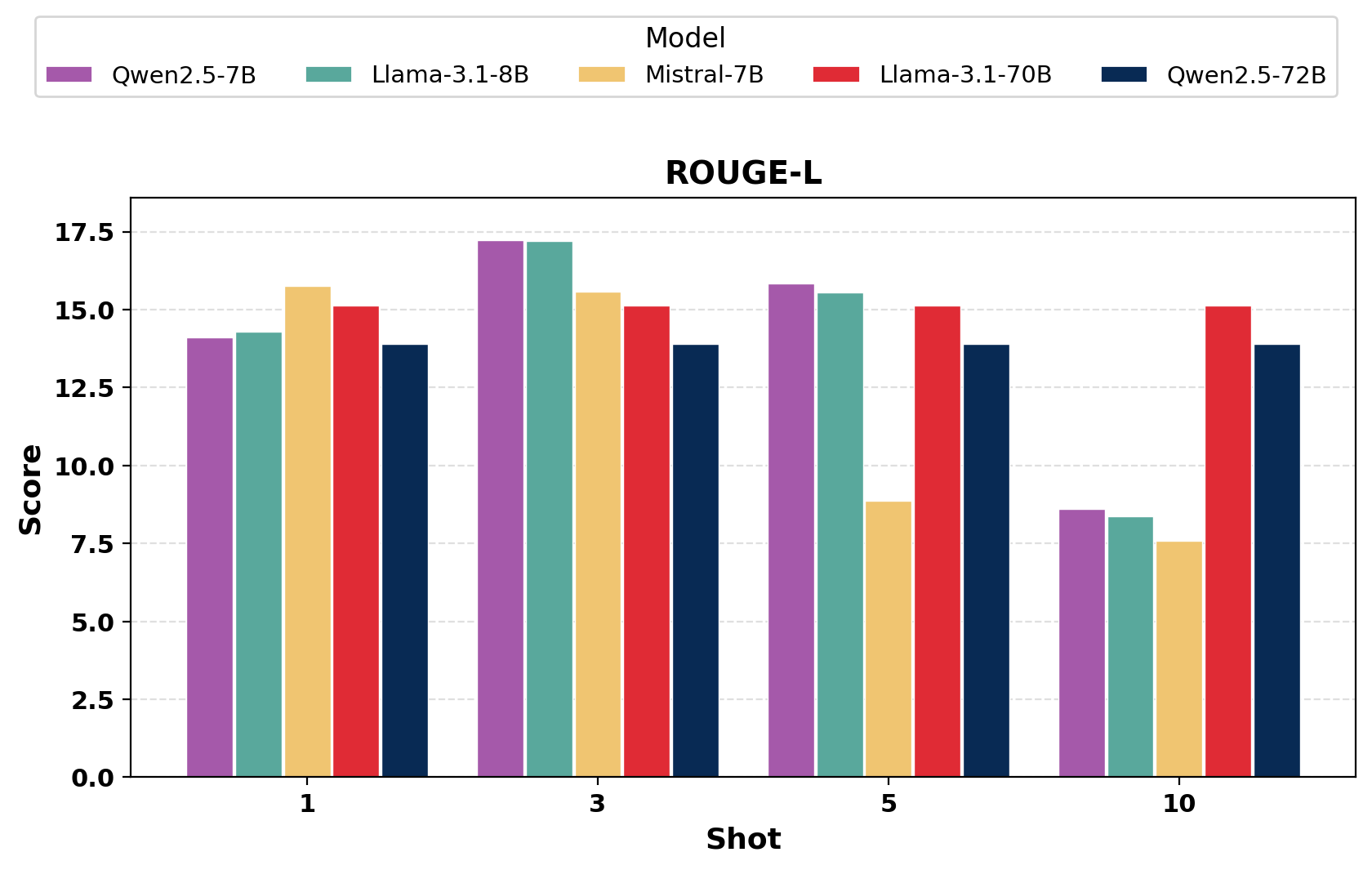}
    \caption{Open-ended ROUGE-L across shot counts (1, 3, 5, 10).}
    \label{fig:open-few-shot-main}
\end{figure}

\vspace{1mm} \noindent \textbf{Few-shot Prompting.}
\label{sec:ablation-fewshot}
Figures~\ref{fig:close-few-shot} and~\ref{fig:open-few-shot-main} show accuracy and ROUGE-L as the number of in-context examples varies over $\{1, 3, 5, 10\}$. Larger models (LLaMA-3.1-70B, Qwen2.5-72B) are essentially unaffected by shot count across all metrics, suggesting they can extract the task pattern from a single example or from zero-shot prompting equally well. In contrast, smaller 7--8B models show sharp degradation at 5 and 10 shots: LLaMA-3.1-8B accuracy collapses from 82.89\% (1-shot) to 10.06\% (10-shot), and ROUGE-L drops from 14.29 to 8.38, as the long few-shot context overwhelms the model's ability to locate the target instruction. Mistral-7B and Qwen2.5-7B follow the same pattern. Notably, 3-shot prompting is the sweet spot for open-ended ROUGE-L: LLaMA-3.1-8B reaches 17.19 at 3 shots (vs.\ 14.29 at 1-shot), and Qwen2.5-7B reaches 17.22, before degrading at higher shot counts. For MCQ accuracy, even 3 shots already reduces performance for most small models, pointing to the inherent tension between providing helpful demonstrations and staying within the model's effective context capacity. Additional open-ended metric trends across all shot counts are shown in Figure~\ref{fig:open-few-shot-appeindex} in the appendix.

\begin{table}[h!]
\centering
\resizebox{\columnwidth}{!}{%
\begin{tabular}{lccccc}
\toprule
\textbf{Model} & \textbf{w/o RAG} & \textbf{BM25} & \textbf{Hybrid} & \textbf{MedCPT} & \textbf{TF-IDF} \\
\midrule
LLaMA3.1-8B  & 0.410 & 0.580 & 0.540 & 0.440 & 0.530 \\
LLaMA3.1-70B & 0.410 & 0.660 & 0.610 & 0.540 & 0.660 \\
Mistral-7B   & 0.460 & 0.510 & 0.490 & 0.390 & 0.530 \\
Qwen2.5-7B   & 0.410 & 0.380 & 0.400 & 0.360 & 0.390 \\
Qwen2.5-72B  & 0.380 & 0.350 & 0.350 & 0.360 & 0.380 \\
\bottomrule
\end{tabular}}
\caption{Accuracy of LLMs across retrieval methods in the oracle retrieval setting, where all retrieved documents are relevant (clean context).}
\label{tab:clean_retrieval}
\end{table}

\vspace{1mm} \noindent \textbf{Quality of the Retrieval Analysis.} Tables~\ref{tab:clean_retrieval} and \ref{tab:noisy_retrieval} show two important problems for retrieval-augmented generation in the biomedical domain. For this analysis, we use the BioASQ corpus as the retrieval source and evaluate on PubMedQA~\cite{jin-2019-pubmedqa}, a benchmark of expert-annotated yes/no/maybe biomedical research questions derived from PubMed abstracts. Since both the retrieval corpus and evaluation dataset come from PubMed, this provides a controlled setting for studying whether retrieved biomedical papers help models answer research questions. To evaluate retrieval quality, we use an LLM-as-a-judge framework to determine whether the retrieved context contains enough information to answer the question correctly. We then select 100 questions where all retrieval methods retrieved context judged to be relevant. The questions are the same across all retrieval methods, but the retrieved documents can differ depending on the retriever.

Table~\ref{tab:clean_retrieval} shows that even when all retrieved contexts contain the correct information, retrieval only leads to limited and inconsistent improvements. For example, LLaMA3.1-70B improves substantially with BM25 retrieval (0.410 $\rightarrow$ 0.660), while Qwen2.5-72B shows almost no improvement across retrieval methods. In several cases, simple sparse retrieval methods such as BM25 and TF-IDF perform better than MedCPT. These results suggest that retrieving relevant evidence alone is not enough to guarantee better performance. Instead, many models still struggle to correctly use and reason over the retrieved information. Table~\ref{tab:noisy_retrieval} further shows that current models are highly sensitive to irrelevant context. When we add 20 unrelated documents to the retrieved evidence, performance drops substantially across nearly all models and retrieval methods. For example, LLaMA3.1-70B decreases from 0.660 to 0.260 under BM25 retrieval, while Mistral-7B drops from 0.530 to 0.340 under TF-IDF retrieval. In many cases, performance becomes worse than using no retrieval at all. Overall, these results show that current biomedical RAG systems remain brittle. Even when relevant evidence is retrieved successfully, small amounts of distracting context can strongly reduce answer accuracy.

\begin{table}[t]
\centering
\resizebox{\columnwidth}{!}{%
\begin{tabular}{lccccc}
\toprule
\textbf{Model} & \textbf{w/o RAG} & \textbf{BM25} & \textbf{Hybrid} & \textbf{MedCPT} & \textbf{TF-IDF} \\
\midrule
LLaMA3.1-8B  & 0.410 & 0.300 & 0.250 & 0.300 & 0.240 \\
LLaMA3.1-70B & 0.410 & 0.260 & 0.260 & 0.340 & 0.290 \\
Mistral-7B   & 0.460 & 0.310 & 0.330 & 0.350 & 0.340 \\
Qwen2.5-7B   & 0.410 & 0.310 & 0.260 & 0.290 & 0.250 \\
Qwen2.5-72B  & 0.380 & 0.250 & 0.240 & 0.280 & 0.230 \\
\bottomrule
\end{tabular}}
\caption{Accuracy of LLMs across retrieval methods in the noisy retrieval setting, where 20 unrelated documents are mixed with retrieved results (distracted context).}
\label{tab:noisy_retrieval}
\end{table}

\vspace{2mm}
\noindent \textbf{Implications.} Our results suggest a more cautious view of biomedical RAG. Retrieval can help, but only when the system retrieves information that is actually relevant to the question. This is not guaranteed, especially when the answer is absent from the corpus or when the retrieved passages are only loosely related. In these cases, retrieval may add little useful information and can introduce misleading context.

Even when relevant evidence is retrieved, the model still has to understand and use it correctly. Our clean retrieval analysis shows that relevant context does not always improve performance, suggesting that evidence use is a major bottleneck. The noisy retrieval results make this concern stronger: adding unrelated documents to useful evidence often hurts performance, sometimes making RAG worse than no retrieval at all. Future biomedical RAG systems, therefore, need better evidence filtering, reranking, and generation methods that can identify useful passages while ignoring distractors.

\section{Conclusion}
\label{sec:conclusion}
We presented a large-scale evaluation of retrieval-augmented generation for biomedical question answering using five open-weight, instruction-tuned models ranging from 7B to 72B parameters. Across all five models, ten datasets, four retrieval methods, and four retrieval corpora, retrieval yields only small and inconsistent improvements over a no-retrieval baseline, typically within 1--2 points on any metric. In contrast, backbone model choice has a substantially larger effect: the gap between a 7B model and its 70B counterpart often exceeds the gain from any retrieval configuration. Expert and layman retrieval corpora also perform similarly in most settings, and differences across retrieval methods (BM25, TF-IDF, MedCPT, Hybrid) remain minor throughout.

Our ablation studies further support this overall pattern. Increasing the number of retrieved documents beyond $k{=}5$ provides little additional benefit for open-ended settings, and few-shot prompting yields a modest gain at 3 shots for smaller models but degrades sharply at higher counts, with small models struggling under longer few-shot contexts. Larger models are comparatively stable across shot counts, but they also show limited benefit from retrieval augmentation.

Taken together, these findings suggest that improving retrieval quality alone may not be sufficient to substantially improve biomedical QA performance in these settings. One possible explanation is that current models, especially smaller ones, do not consistently make effective use of retrieved evidence, though our experiments do not directly measure evidence utilization or grounding. This points to several directions for future work, including training or fine-tuning methods that better support evidence integration, post-retrieval re-ranking or filtering to reduce context noise, and evaluation frameworks that more directly assess faithfulness and grounding rather than relying only on reference-based metrics. More broadly, an important open question is when retrieval is actually necessary, and whether we can better identify cases where the required knowledge is already contained within the model.

\section*{Acknowledgments}
This material is based upon work supported by the National Science Foundation (NSF) under Grant~No. 2145357.

\section*{Limitations}

Our study has several limitations. {First, we evaluate retrieval-augmented generation using only reference-based downstream metrics such as ROUGE-L, BLEU, METEOR, BERTScore, and accuracy, rather than direct measures of faithfulness or evidence grounding.} For example, a model may produce a correct answer from its parametric knowledge without actually using the retrieved documents, or it may copy surface details from retrieval without truly improving medical correctness. This limitation is not major for our study because our main goal is a comparative evaluation across models, retrievers, corpora, and no-retrieval baselines within a single, consistent framework, and these metrics are sufficient to support our central finding that retrieval provides only small and inconsistent gains.

{Second, our experiments are limited to five open-weight instruction-tuned models and do not include proprietary frontier systems such as GPT-4-class medical assistants.} It is possible that stronger closed models use retrieved evidence more effectively, especially in cases requiring multi-step reasoning over documents. This limitation is not major for our study because our paper is specifically motivated by practical, deployable biomedical QA settings, where open-weight 7B--72B models are realistic choices, and we also include both small and large open models to test whether the observed pattern holds across scales.

{Third, our retrieval setup uses a fixed top-$k$ pipeline with four retrievers and four corpora, but does not explore more complex retrieval strategies such as adaptive retrieval, document re-ranking, iterative retrieval, or task-specific chunking.} For instance, some questions may require retrieving fewer but more precise passages, while others may benefit from multi-hop retrieval or filtering noisy evidence before generation. This limitation is not major for our study because we intentionally focus on strong, standard retrieval baselines widely used in prior biomedical RAG work, which makes the comparison clean and allows us to show that, even with several commonly used retrieval choices, the gains remain modest.

{Finally, our evaluation mixes expert and layman biomedical QA datasets, but the study does not separately analyze all possible sources of variation across question type, answer length, or knowledge intensity.} For example, retrieval may be more useful for highly specialized factoid questions than for common consumer-health questions that models may already answer from pretraining alone. This limitation is not major for our study because the breadth of datasets is a strength of the paper overall: the consistency of the pattern across ten datasets suggests that the weak benefit of retrieval is not tied to a single benchmark or user population.

\bibliography{custom}

@inproceedings{abacha2019summarization,
    title = "On the Summarization of Consumer Health Questions",
    author = "Ben Abacha, Asma  and
      Demner-Fushman, Dina",
    editor = "Korhonen, Anna  and
      Traum, David  and
      M{\`a}rquez, Llu{\'i}s",
    booktitle = "Proceedings of the 57th Annual Meeting of the Association for Computational Linguistics",
    month = jul,
    year = "2019",
    address = "Florence, Italy",
    publisher = "Association for Computational Linguistics",
    url = "https://aclanthology.org/P19-1215/",
    doi = "10.18653/v1/P19-1215",
    pages = "2228--2234"
}

@conference {nentidis2025overview,
	title = {Overview of~BioASQ 2025: The Thirteenth BioASQ Challenge on~Large-Scale Biomedical Semantic Indexing and~Question Answering},
	booktitle = {Experimental IR Meets Multilinguality, Multimodality, and Interaction},
	year = {2025},
	publisher = {Springer Nature Switzerland},
	organization = {Springer Nature Switzerland},
	address = {Cham},
	isbn = {978-3-032-04354-2},
	doi = {10.1007/978-3-032-04354-2_12},
	url = {https://link.springer.com/10.1007/978-3-032-04354-2_12},
	author = {Nentidis, Anastasios and Katsimpras, Georgios and Krithara, Anastasia and Krallinger, Martin and Rodr{\'\i}guez-Ortega, Miguel and Rodriguez-L{\'o}pez, Eduard and Loukachevitch, Natalia and Sakhovskiy, Andrey and Tutubalina, Elena and Dimitriadis, Dimitris and Tsoumakas, Grigorios and Giannakoulas, George and Bekiaridou, Alexandra and Samaras, Athanasios and Nunzio, Giorgio Maria Di and Ferro, Nicola and Marchesin, Stefano and Martinelli, Marco and Silvello, Gianmaria and Paliouras, Georgios}
}

@inproceedings{nguyen2023medredqa,
    title = "{M}ed{R}ed{QA} for Medical Consumer Question Answering: Dataset, Tasks, and Neural Baselines",
    author = "Nguyen, Vincent  and
      Karimi, Sarvnaz  and
      Rybinski, Maciej  and
      Xing, Zhenchang",
    editor = "Park, Jong C.  and
      Arase, Yuki  and
      Hu, Baotian  and
      Lu, Wei  and
      Wijaya, Derry  and
      Purwarianti, Ayu  and
      Krisnadhi, Adila Alfa",
    booktitle = "Proceedings of the 13th International Joint Conference on Natural Language Processing and the 3rd Conference of the Asia-Pacific Chapter of the Association for Computational Linguistics (Volume 1: Long Papers)",
    month = nov,
    year = "2023",
    address = "Nusa Dua, Bali",
    publisher = "Association for Computational Linguistics",
    url = "https://aclanthology.org/2023.ijcnlp-main.42/",
    doi = "10.18653/v1/2023.ijcnlp-main.42",
    pages = "629--648"
}

@article{ben2019question,
   author = {Ben Abacha, Asma and Demner-Fushman, Dina},
   title = {A question-entailment approach to question answering},
   journal = {BMC Bioinformatics},
   volume = {20},
   number = {1},
   pages = {511},
   ISSN = {1471-2105},
   DOI = {10.1186/s12859-019-3119-4},
   url = {https://doi.org/10.1186/s12859-019-3119-4},
   year = {2019},
   type = {Journal Article}
}

@article{abacha2019bridging,
   author = {Abacha, A. B. and Mrabet, Y. and Sharp, M. and Goodwin, T. R. and Shooshan, S. E. and Demner-Fushman, D.},
   title = {Bridging the Gap Between Consumers' Medication Questions and Trusted Answers},
   journal = {Stud Health Technol Inform},
   volume = {264},
   pages = {25-29},
   note = {1879-8365
Abacha, Asma Ben
Mrabet, Yassine
Sharp, Mark
Goodwin, Travis R
Shooshan, Sonya E
Demner-Fushman, Dina
Journal Article
Netherlands
2019/08/24
Stud Health Technol Inform. 2019 Aug 21;264:25-29. doi: 10.3233/SHTI190176.},
   ISSN = {0926-9630},
   DOI = {10.3233/shti190176},
   year = {2019},
   type = {Journal Article}
}

@article{jin2021disease,
AUTHOR = {Jin, Di and Pan, Eileen and Oufattole, Nassim and Weng, Wei-Hung and Fang, Hanyi and Szolovits, Peter},
TITLE = {What Disease Does This Patient Have? A Large-Scale Open Domain Question Answering Dataset from Medical Exams},
JOURNAL = {Applied Sciences},
VOLUME = {11},
YEAR = {2021},
NUMBER = {14},
ARTICLE-NUMBER = {6421},
URL = {https://www.mdpi.com/2076-3417/11/14/6421},
ISSN = {2076-3417},
DOI = {10.3390/app11146421}
}

@inproceedings{zhu2020question,
    title = "Question Answering with Long Multiple-Span Answers",
    author = "Zhu, Ming  and
      Ahuja, Aman  and
      Juan, Da-Cheng  and
      Wei, Wei  and
      Reddy, Chandan K.",
    editor = "Cohn, Trevor  and
      He, Yulan  and
      Liu, Yang",
    booktitle = "Findings of the Association for Computational Linguistics: EMNLP 2020",
    month = nov,
    year = "2020",
    address = "Online",
    publisher = "Association for Computational Linguistics",
    url = "https://aclanthology.org/2020.findings-emnlp.342/",
    doi = "10.18653/v1/2020.findings-emnlp.342",
    pages = "3840--3849"
}

@InProceedings{pal2022medmcqa,
  title = 	 {MedMCQA: A Large-scale Multi-Subject Multi-Choice Dataset for Medical domain Question Answering},
  author =       {Pal, Ankit and Umapathi, Logesh Kumar and Sankarasubbu, Malaikannan},
  booktitle = 	 {Proceedings of the Conference on Health, Inference, and Learning},
  pages = 	 {248--260},
  year = 	 {2022},
  editor = 	 {Flores, Gerardo and Chen, George H and Pollard, Tom and Ho, Joyce C and Naumann, Tristan},
  volume = 	 {174},
  series = 	 {Proceedings of Machine Learning Research},
  month = 	 {07--08 Apr},
  publisher =    {PMLR},
  url = 	 {https://proceedings.mlr.press/v174/pal22a.html}
}

@article{li2023chatdoctor,
  title={Chatdoctor: A medical chat model fine-tuned on a large language model meta-ai (llama) using medical domain knowledge},
  author={Li, Yunxiang and Li, Zihan and Zhang, Kai and Dan, Ruilong and Jiang, Steve and Zhang, You},
  journal={Cureus},
  volume={15},
  number={6},
  year={2023},
  publisher={Cureus},
  url={https://pmc.ncbi.nlm.nih.gov/articles/PMC10364849/}
}

@inproceedings{
hendrycks2020measuring,
title={Measuring Massive Multitask Language Understanding},
author={Dan Hendrycks and Collin Burns and Steven Basart and Andy Zou and Mantas Mazeika and Dawn Song and Jacob Steinhardt},
booktitle={International Conference on Learning Representations},
year={2021},
url={https://openreview.net/forum?id=d7KBjmI3GmQ}
}

@article{krithara2023bioasq,
   author = {Krithara, A. and Nentidis, A. and Bougiatiotis, K. and Paliouras, G.},
   title = {BioASQ-QA: A manually curated corpus for Biomedical Question Answering},
   journal = {Sci Data},
   volume = {10},
   number = {1},
   pages = {170},
   note = {2052-4463
Krithara, Anastasia
Orcid: 0000-0003-0491-4507
Nentidis, Anastasios
Bougiatiotis, Konstantinos
Paliouras, Georgios
Dataset
Journal Article
England
2023/03/28
Sci Data. 2023 Mar 27;10(1):170. doi: 10.1038/s41597-023-02068-4.},
   ISSN = {2052-4463},
   DOI = {10.1038/s41597-023-02068-4},
   year = {2023},
   type = {Journal Article}
}

@techreport{yahoo_webscope_2009,
  title        = {{Yahoo! Webscope Datasets Catalog}},
  author       = {{Yahoo! Research}},
  institution  = {Yahoo! Inc.},
  year         = {2009},
  month        = jan,
  note         = {19 Datasets Available. Accessed via Stanford InfoLab},
  url          = {http://infolab.stanford.edu/~ullman/mining/2009/YahooData.pdf}
}

@inproceedings{xiong2024benchmarking,
    title = "Benchmarking Retrieval-Augmented Generation for Medicine",
    author = "Xiong, Guangzhi  and
      Jin, Qiao  and
      Lu, Zhiyong  and
      Zhang, Aidong",
    editor = "Ku, Lun-Wei  and
      Martins, Andre  and
      Srikumar, Vivek",
    booktitle = "Findings of the Association for Computational Linguistics: ACL 2024",
    month = aug,
    year = "2024",
    address = "Bangkok, Thailand",
    publisher = "Association for Computational Linguistics",
    url = "https://aclanthology.org/2024.findings-acl.372/",
    doi = "10.18653/v1/2024.findings-acl.372",
    pages = "6233--6251"
}

@inproceedings{zhang2022focus,
    title = "Focus-Driven Contrastive Learning for Medical Question Summarization",
    author = "Zhang, Ming  and
      Dou, Shuai  and
      Wang, Ziyang  and
      Wu, Yunfang",
    editor = "Calzolari, Nicoletta  and
      Huang, Chu-Ren  and
      Kim, Hansaem  and
      Pustejovsky, James  and
      Wanner, Leo  and
      Choi, Key-Sun  and
      Ryu, Pum-Mo  and
      Chen, Hsin-Hsi  and
      Donatelli, Lucia  and
      Ji, Heng  and
      Kurohashi, Sadao  and
      Paggio, Patrizia  and
      Xue, Nianwen  and
      Kim, Seokhwan  and
      Hahm, Younggyun  and
      He, Zhong  and
      Lee, Tony Kyungil  and
      Santus, Enrico  and
      Bond, Francis  and
      Na, Seung-Hoon",
    booktitle = "Proceedings of the 29th International Conference on Computational Linguistics",
    month = oct,
    year = "2022",
    address = "Gyeongju, Republic of Korea",
    publisher = "International Committee on Computational Linguistics",
    url = "https://aclanthology.org/2022.coling-1.539/",
    pages = "6176--6186"
}

@inproceedings{tang2024medagents,
    title = "{M}ed{A}gents: Large Language Models as Collaborators for Zero-shot Medical Reasoning",
    author = "Tang, Xiangru  and
      Zou, Anni  and
      Zhang, Zhuosheng  and
      Li, Ziming  and
      Zhao, Yilun  and
      Zhang, Xingyao  and
      Cohan, Arman  and
      Gerstein, Mark",
    editor = "Ku, Lun-Wei  and
      Martins, Andre  and
      Srikumar, Vivek",
    booktitle = "Findings of the Association for Computational Linguistics: ACL 2024",
    month = aug,
    year = "2024",
    address = "Bangkok, Thailand",
    publisher = "Association for Computational Linguistics",
    url = "https://aclanthology.org/2024.findings-acl.33/",
    doi = "10.18653/v1/2024.findings-acl.33",
    pages = "599--621"
}

@article{robertson2009probabilistic,
author = {Robertson, Stephen and Zaragoza, Hugo},
title = {The Probabilistic Relevance Framework: BM25 and Beyond},
year = {2009},
issue_date = {April 2009},
publisher = {Now Publishers Inc.},
address = {Hanover, MA, USA},
volume = {3},
number = {4},
issn = {1554-0669},
url = {https://doi.org/10.1561/1500000019},
doi = {10.1561/1500000019},
journal = {Found. Trends Inf. Retr.},
month = apr,
pages = {333–389},
numpages = {57}
}

@article{sparck1972statistical,
    author = {SPARCK JONES, KAREN},
    title = {A STATISTICAL INTERPRETATION OF TERM SPECIFICITY AND ITS APPLICATION IN RETRIEVAL},
    journal = {Journal of Documentation},
    volume = {28},
    number = {1},
    pages = {11-21},
    year = {1972},
    month = {01},
    issn = {0022-0418},
    doi = {10.1108/eb026526},
    url = {https://doi.org/10.1108/eb026526}
}

@article{jin2023medcpt,
    author = {Jin, Qiao and Kim, Won and Chen, Qingyu and Comeau, Donald C and Yeganova, Lana and Wilbur, W John and Lu, Zhiyong},
    title = {MedCPT: Contrastive Pre-trained Transformers with large-scale PubMed search logs for zero-shot biomedical information retrieval},
    journal = {Bioinformatics},
    volume = {39},
    number = {11},
    pages = {btad651},
    year = {2023},
    month = {11},
    issn = {1367-4811},
    doi = {10.1093/bioinformatics/btad651},
    url = {https://doi.org/10.1093/bioinformatics/btad651},
    eprint = {https://academic.oup.com/bioinformatics/article-pdf/39/11/btad651/52799559/btad651.pdf},
}

@inproceedings{cormack2009reciprocal,
author = {Cormack, Gordon V. and Clarke, Charles L A and Buettcher, Stefan},
title = {Reciprocal rank fusion outperforms condorcet and individual rank learning methods},
year = {2009},
isbn = {9781605584836},
publisher = {Association for Computing Machinery},
address = {New York, NY, USA},
url = {https://doi.org/10.1145/1571941.1572114},
doi = {10.1145/1571941.1572114},
booktitle = {Proceedings of the 32nd International ACM SIGIR Conference on Research and Development in Information Retrieval},
pages = {758–759},
numpages = {2},
location = {Boston, MA, USA},
series = {SIGIR '09}
}

@article{yang2025qwen3,
  title={Qwen3 technical report},
  author={Yang, An and Li, Anfeng and Yang, Baosong and Zhang, Beichen and Hui, Binyuan and Zheng, Bo and Yu, Bowen and Gao, Chang and Huang, Chengen and Lv, Chenxu and Zheng, Chujie and Liu, Dayiheng and Zhou, Fan and Huang, Fei and Hu, Feng and Ge, Hao and Wei, Haoran and Lin, Huan and Tang, Jialong and Yang, Jian and Tu, Jianhong and Zhang, Jianwei and Yang, Jianxin and Yang, Jiaxi and Zhou, Jing and Zhou, Jingren and Lin, Junyang and Dang, Kai and Bao, Keqin and Yang, Kexin and Yu, Le and Deng, Lianghao and Li, Mei and Xue, Mingfeng and Li, Mingze and Zhang, Pei and Wang, Peng and Zhu, Qin and Men, Rui and Gao, Ruize and Liu, Shixuan and Luo, Shuang and Li, Tianhao and Tang, Tianyi and Yin, Wenbiao and Ren, Xingzhang and Wang, Xinyu and Zhang, Xinyu and Ren, Xuancheng and Fan, Yang and Su, Yang and Zhang, Yichang and Zhang, Yinger and Wan, Yu and Liu, Yuqiong and Wang, Zekun and Cui, Zeyu and Zhang, Zhenru and Zhou, Zhipeng and Qiu, Zihan},
  journal={arXiv preprint arXiv:2505.09388},
  year={2025},
  url={https://arxiv.org/abs/2505.09388}
}

@ARTICLE{grattafiori2024llama,
       author = {{Grattafiori}, Aaron and {Dubey}, Abhimanyu and {Jauhri}, Abhinav and {Pandey}, Abhinav and {Kadian}, Abhishek and {Al-Dahle}, Ahmad and {Letman}, Aiesha and {Mathur}, Akhil and {Schelten}, Alan and {Vaughan}, Alex and {Yang}, Amy and {Fan}, Angela and {Goyal}, Anirudh and {Hartshorn}, Anthony and {Yang}, Aobo and {Mitra}, Archi and {Sravankumar}, Archie and {Korenev}, Artem and {Hinsvark}, Arthur and {Rao}, Arun and {Zhang}, Aston and {Rodriguez}, Aurelien and {Gregerson}, Austen and {Spataru}, Ava and {Roziere}, Baptiste and {Biron}, Bethany and {Tang}, Binh and {Chern}, Bobbie and {Caucheteux}, Charlotte and {Nayak}, Chaya and {Bi}, Chloe and {Marra}, Chris and {McConnell}, Chris and {Keller}, Christian and {Touret}, Christophe and {Wu}, Chunyang and {Wong}, Corinne and {Canton Ferrer}, Cristian and {Nikolaidis}, Cyrus and {Allonsius}, Damien and {Song}, Daniel and {Pintz}, Danielle and {Livshits}, Danny and {Wyatt}, Danny and {Esiobu}, David and {Choudhary}, Dhruv and {Mahajan}, Dhruv and {Garcia-Olano}, Diego and {Perino}, Diego and {Hupkes}, Dieuwke and {Lakomkin}, Egor and {AlBadawy}, Ehab and {Lobanova}, Elina and {Dinan}, Emily and {Smith}, Eric Michael and {Radenovic}, Filip and {Guzm{\'a}n}, Francisco and {Zhang}, Frank and {Synnaeve}, Gabriel and {Lee}, Gabrielle and {Anderson}, Georgia Lewis and {Thattai}, Govind and {Nail}, Graeme and {Mialon}, Gregoire and {Pang}, Guan and {Cucurell}, Guillem and {Nguyen}, Hailey and {Korevaar}, Hannah and {Xu}, Hu and {Touvron}, Hugo and {Zarov}, Iliyan and {Arrieta Ibarra}, Imanol and {Kloumann}, Isabel and {Misra}, Ishan and {Evtimov}, Ivan and {Zhang}, Jack and {Copet}, Jade and {Lee}, Jaewon and {Geffert}, Jan and {Vranes}, Jana and {Park}, Jason and {Mahadeokar}, Jay and {Shah}, Jeet and {van der Linde}, Jelmer and {Billock}, Jennifer and {Hong}, Jenny and {Lee}, Jenya and {Fu}, Jeremy and {Chi}, Jianfeng and {Huang}, Jianyu and {Liu}, Jiawen and {Wang}, Jie and {Yu}, Jiecao and {Bitton}, Joanna and {Spisak}, Joe and {Park}, Jongsoo and {Rocca}, Joseph and {Johnstun}, Joshua and {Saxe}, Joshua and {Jia}, Junteng and {Vasuden Alwala}, Kalyan and {Prasad}, Karthik and {Upasani}, Kartikeya and {Plawiak}, Kate and {Li}, Ke and {Heafield}, Kenneth and {Stone}, Kevin and {El-Arini}, Khalid and {Iyer}, Krithika and {Malik}, Kshitiz and {Chiu}, Kuenley and {Bhalla}, Kunal and {Lakhotia}, Kushal and {Rantala-Yeary}, Lauren and {van der Maaten}, Laurens and {Chen}, Lawrence and {Tan}, Liang and {Jenkins}, Liz and {Martin}, Louis and {Madaan}, Lovish and {Malo}, Lubo and {Blecher}, Lukas and {Landzaat}, Lukas and {de Oliveira}, Luke and {Muzzi}, Madeline and {Pasupuleti}, Mahesh and {Singh}, Mannat and {Paluri}, Manohar and {Kardas}, Marcin and {Tsimpoukelli}, Maria and {Oldham}, Mathew and {Rita}, Mathieu and {Pavlova}, Maya and {Kambadur}, Melanie and {Lewis}, Mike and {Si}, Min and {Singh}, Mitesh Kumar and {Hassan}, Mona and {Goyal}, Naman and {Torabi}, Narjes and {Bashlykov}, Nikolay and {Bogoychev}, Nikolay and {Chatterji}, Niladri and {Zhang}, Ning and {Duchenne}, Olivier and {{\c{C}}elebi}, Onur and {Alrassy}, Patrick and {Zhang}, Pengchuan and {Li}, Pengwei and {Vasic}, Petar and {Weng}, Peter and {Bhargava}, Prajjwal and {Dubal}, Pratik and {Krishnan}, Praveen and {Singh Koura}, Punit and {Xu}, Puxin and {He}, Qing and {Dong}, Qingxiao and {Srinivasan}, Ragavan and {Ganapathy}, Raj and {Calderer}, Ramon and {Silveira Cabral}, Ricardo and {Stojnic}, Robert and {Raileanu}, Roberta and {Maheswari}, Rohan and {Girdhar}, Rohit and {Patel}, Rohit and {Sauvestre}, Romain and {Polidoro}, Ronnie and {Sumbaly}, Roshan and {Taylor}, Ross and {Silva}, Ruan and {Hou}, Rui and {Wang}, Rui and {Hosseini}, Saghar and {Chennabasappa}, Sahana and {Singh}, Sanjay and {Bell}, Sean and {Kim}, Seohyun Sonia and {Edunov}, Sergey and {Nie}, Shaoliang and {Narang}, Sharan and {Raparthy}, Sharath and {Shen}, Sheng and {Wan}, Shengye and {Bhosale}, Shruti and {Zhang}, Shun and {Vandenhende}, Simon and {Batra}, Soumya and {Whitman}, Spencer and {Sootla}, Sten and {Collot}, Stephane and {Gururangan}, Suchin and {Borodinsky}, Sydney and {Herman}, Tamar and {Fowler}, Tara and {Sheasha}, Tarek and {Georgiou}, Thomas and {Scialom}, Thomas and {Speckbacher}, Tobias},
        title = "{The Llama 3 Herd of Models}",
      journal = {arXiv e-prints},
         year = 2024,
        month = jul,
          eid = {arXiv:2407.21783},
        pages = {arXiv:2407.21783},
          doi = {10.48550/arXiv.2407.21783},
archivePrefix = {arXiv},
       eprint = {2407.21783},
 primaryClass = {cs.AI}
}

@article{jiang20236g,
  title={Mistral 7B},
  author={Jiang, Albert Q and Sablayrolles, Alexandre and Mensch, Arthur and Bamford, Chris and Chaplot, Devendra Singh and de las Casas, Diego and Bressand, Florian and Lengyel, Gianna and Lample, Guillaume and Saulnier, Lucile and others},
  journal={arXiv preprint arXiv:2310.06825},
  year={2023},
  url={https://arxiv.org/abs/2310.06825}
}

@inproceedings{lewis2020retrieval,
 author = {Lewis, Patrick and Perez, Ethan and Piktus, Aleksandra and Petroni, Fabio and Karpukhin, Vladimir and Goyal, Naman and K\"{u}ttler, Heinrich and Lewis, Mike and Yih, Wen-tau and Rockt\"{a}schel, Tim and Riedel, Sebastian and Kiela, Douwe},
 booktitle = {Advances in Neural Information Processing Systems},
 editor = {H. Larochelle and M. Ranzato and R. Hadsell and M.F. Balcan and H. Lin},
 pages = {9459--9474},
 publisher = {Curran Associates, Inc.},
 title = {Retrieval-Augmented Generation for Knowledge-Intensive NLP Tasks},
 url = {https://proceedings.neurips.cc/paper_files/paper/2020/file/6b493230205f780e1bc26945df7481e5-Paper.pdf},
 volume = {33},
 year = {2020}
}

@article{singhal2023large,
   author = {Singhal, Karan and Azizi, Shekoofeh and Tu, Tao and Mahdavi, S. Sara and Wei, Jason and Chung, Hyung Won and Scales, Nathan and Tanwani, Ajay and Cole-Lewis, Heather and Pfohl, Stephen and Payne, Perry and Seneviratne, Martin and Gamble, Paul and Kelly, Chris and Babiker, Abubakr and Schärli, Nathanael and Chowdhery, Aakanksha and Mansfield, Philip and Demner-Fushman, Dina and Agüera y Arcas, Blaise and Webster, Dale and Corrado, Greg S. and Matias, Yossi and Chou, Katherine and Gottweis, Juraj and Tomasev, Nenad and Liu, Yun and Rajkomar, Alvin and Barral, Joelle and Semturs, Christopher and Karthikesalingam, Alan and Natarajan, Vivek},
   title = {Large language models encode clinical knowledge},
   journal = {Nature},
   volume = {620},
   number = {7972},
   pages = {172-180},
   ISSN = {1476-4687},
   DOI = {10.1038/s41586-023-06291-2},
   url = {https://doi.org/10.1038/s41586-023-06291-2},
   year = {2023},
   type = {Journal Article}
}

@article{ji2023survey,
author = {Ji, Ziwei and Lee, Nayeon and Frieske, Rita and Yu, Tiezheng and Su, Dan and Xu, Yan and Ishii, Etsuko and Bang, Ye Jin and Madotto, Andrea and Fung, Pascale},
title = {Survey of Hallucination in Natural Language Generation},
year = {2023},
issue_date = {December 2023},
publisher = {Association for Computing Machinery},
address = {New York, NY, USA},
volume = {55},
number = {12},
issn = {0360-0300},
url = {https://doi.org/10.1145/3571730},
doi = {10.1145/3571730},
journal = {ACM Comput. Surv.},
month = mar,
articleno = {248},
numpages = {38}
}

@inproceedings{karpukhin2020dense,
    title = "Dense Passage Retrieval for Open-Domain Question Answering",
    author = "Karpukhin, Vladimir  and
      Oguz, Barlas  and
      Min, Sewon  and
      Lewis, Patrick  and
      Wu, Ledell  and
      Edunov, Sergey  and
      Chen, Danqi  and
      Yih, Wen-tau",
    editor = "Webber, Bonnie  and
      Cohn, Trevor  and
      He, Yulan  and
      Liu, Yang",
    booktitle = "Proceedings of the 2020 Conference on Empirical Methods in Natural Language Processing (EMNLP)",
    month = nov,
    year = "2020",
    address = "Online",
    publisher = "Association for Computational Linguistics",
    url = "https://aclanthology.org/2020.emnlp-main.550/",
    doi = "10.18653/v1/2020.emnlp-main.550",
    pages = "6769--6781"
}

@inproceedings{trivedi2023interleaving,
    title = "Interleaving Retrieval with Chain-of-Thought Reasoning for Knowledge-Intensive Multi-Step Questions",
    author = "Trivedi, Harsh  and
      Balasubramanian, Niranjan  and
      Khot, Tushar  and
      Sabharwal, Ashish",
    editor = "Rogers, Anna  and
      Boyd-Graber, Jordan  and
      Okazaki, Naoaki",
    booktitle = "Proceedings of the 61st Annual Meeting of the Association for Computational Linguistics (Volume 1: Long Papers)",
    month = jul,
    year = "2023",
    address = "Toronto, Canada",
    publisher = "Association for Computational Linguistics",
    url = "https://aclanthology.org/2023.acl-long.557/",
    doi = "10.18653/v1/2023.acl-long.557",
    pages = "10014--10037"
}

@inproceedings{shao2023enhancing,
    title = "Enhancing Retrieval-Augmented Large Language Models with Iterative Retrieval-Generation Synergy",
    author = "Shao, Zhihong  and
      Gong, Yeyun  and
      Shen, Yelong  and
      Huang, Minlie  and
      Duan, Nan  and
      Chen, Weizhu",
    editor = "Bouamor, Houda  and
      Pino, Juan  and
      Bali, Kalika",
    booktitle = "Findings of the Association for Computational Linguistics: EMNLP 2023",
    month = dec,
    year = "2023",
    address = "Singapore",
    publisher = "Association for Computational Linguistics",
    url = "https://aclanthology.org/2023.findings-emnlp.620/",
    doi = "10.18653/v1/2023.findings-emnlp.620",
    pages = "9248--9274"
}

@inproceedings{
asai2023self,
title={Self-{RAG}: Learning to Retrieve, Generate, and Critique through Self-Reflection},
author={Akari Asai and Zeqiu Wu and Yizhong Wang and Avirup Sil and Hannaneh Hajishirzi},
booktitle={The Twelfth International Conference on Learning Representations},
year={2024},
url={https://openreview.net/forum?id=hSyW5go0v8}
}

@inproceedings{ma2023query,
    title = "Query Rewriting in Retrieval-Augmented Large Language Models",
    author = "Ma, Xinbei  and
      Gong, Yeyun  and
      He, Pengcheng  and
      Zhao, Hai  and
      Duan, Nan",
    editor = "Bouamor, Houda  and
      Pino, Juan  and
      Bali, Kalika",
    booktitle = "Proceedings of the 2023 Conference on Empirical Methods in Natural Language Processing",
    month = dec,
    year = "2023",
    address = "Singapore",
    publisher = "Association for Computational Linguistics",
    url = "https://aclanthology.org/2023.emnlp-main.322/",
    doi = "10.18653/v1/2023.emnlp-main.322",
    pages = "5303--5315"
}

@article{gao2023retrieval,
  title={Retrieval-augmented generation for large language models: A survey},
  author={Gao, Yunfan and Xiong, Yun and Gao, Xinyu and Jia, Kangxiang and Pan, Jinliu and Bi, Yuxi and Dai, Yixin and Sun, Jiawei and Wang, Haofen and Wang, Haofen},
  journal={arXiv preprint arXiv:2312.10997},
  volume={2},
  number={1},
  pages={32},
  year={2023},
  url={https://arxiv.org/abs/2312.10997}
}

@article{nori2023capabilities,
  title={Capabilities of gpt-4 on medical challenge problems},
  author={Nori, Harsha and King, Nicholas and McKinney, Scott Mayer and Carignan, Dean and Horvitz, Eric},
  journal={arXiv preprint arXiv:2303.13375},
  year={2023},
  url={https://arxiv.org/abs/2303.13375}
}

@InProceedings{shi2023large,
  title = 	 {Large Language Models Can Be Easily Distracted by Irrelevant Context},
  author =       {Shi, Freda and Chen, Xinyun and Misra, Kanishka and Scales, Nathan and Dohan, David and Chi, Ed H. and Sch\"{a}rli, Nathanael and Zhou, Denny},
  booktitle = 	 {Proceedings of the 40th International Conference on Machine Learning},
  pages = 	 {31210--31227},
  year = 	 {2023},
  editor = 	 {Krause, Andreas and Brunskill, Emma and Cho, Kyunghyun and Engelhardt, Barbara and Sabato, Sivan and Scarlett, Jonathan},
  volume = 	 {202},
  series = 	 {Proceedings of Machine Learning Research},
  month = 	 {23--29 Jul},
  publisher =    {PMLR},
  url = 	 {https://proceedings.mlr.press/v202/shi23a.html}
}

@inproceedings{ovadia2024fine,
    title = "Fine-Tuning or Retrieval? Comparing Knowledge Injection in {LLM}s",
    author = "Ovadia, Oded  and
      Brief, Menachem  and
      Mishaeli, Moshik  and
      Elisha, Oren",
    editor = "Al-Onaizan, Yaser  and
      Bansal, Mohit  and
      Chen, Yun-Nung",
    booktitle = "Proceedings of the 2024 Conference on Empirical Methods in Natural Language Processing",
    month = nov,
    year = "2024",
    address = "Miami, Florida, USA",
    publisher = "Association for Computational Linguistics",
    url = "https://aclanthology.org/2024.emnlp-main.15/",
    doi = "10.18653/v1/2024.emnlp-main.15",
    pages = "237--250"
}

@inproceedings{jin-2019-pubmedqa,
    title = "{P}ub{M}ed{QA}: A Dataset for Biomedical Research Question Answering",
    author = "Jin, Qiao  and
      Dhingra, Bhuwan  and
      Liu, Zhengping  and
      Cohen, William  and
      Lu, Xinghua",
    editor = "Inui, Kentaro  and
      Jiang, Jing  and
      Ng, Vincent  and
      Wan, Xiaojun",
    booktitle = "Proceedings of the 2019 Conference on Empirical Methods in Natural Language Processing and the 9th International Joint Conference on Natural Language Processing (EMNLP-IJCNLP)",
    month = nov,
    year = "2019",
    address = "Hong Kong, China",
    publisher = "Association for Computational Linguistics",
    url = "https://aclanthology.org/D19-1259/",
    doi = "10.18653/v1/D19-1259",
    pages = "2567--2577"
}

\appendix

\section{Datasets}
\label{sec:appendix-datasets}

Our experiments use ten biomedical and consumer-health query datasets and four retrieval corpora. Table~\ref{tab:query-datasets} summarises each evaluation dataset, its source size, and the number of examples selected for evaluation; Table~\ref{tab:retrieval-corpora} describes the knowledge bases indexed for retrieval. Full details on dataset splits and the few-shot query pools are provided in Section~\ref{sec:datasets-main}.

\begin{table*}[t]
\centering
\scriptsize
\setlength{\tabcolsep}{3.5pt}
\renewcommand{\arraystretch}{1.15}
\newcolumntype{M}[1]{>{\centering\arraybackslash}m{#1}}
\begin{tabular}{M{1.4cm} M{3.0cm} M{1.8cm} M{4.8cm} M{1.1cm} M{1.0cm}}
\toprule
\textbf{User Type} & \textbf{Dataset} & \textbf{Dataset Type} & \textbf{Dataset Content} & \textbf{Query Set} & \textbf{Test Set} \\
\midrule
\multirow{5}{*}{\parbox[c][25em][c]{\linewidth}{\centering Layman}}
  & MeQSum~\cite{abacha2019summarization}
  & Open-ended Q\&A
  & 1,000 consumer health questions from the U.S.\ National Library of Medicine, manually summarized by medical experts (inter-annotator agreement: 96.9\%)
  & 500 & 500 \\
\cmidrule(lr){2-6}
  & MedRedQA~\cite{nguyen2023medredqa}
  & Open-ended Q\&A
  & 51,000 consumer question--physician answer pairs from Reddit \texttt{/r/AskDocs} (2013--2022); answers from verified doctors only; \textasciitilde1k entries enriched with PubMed evidence
  & 45,863 & 1,000 \\
\cmidrule(lr){2-6}
  & MedicationQA~\cite{abacha2019bridging}
  & Open-ended Q\&A
  & 690 real consumer medication questions annotated with drug focus and question type (dosage, side effects, interactions, etc.); answers sourced from MedlinePlus, DailyMed, FDA, and Mayo Clinic
  & 189 & 500 \\
\cmidrule(lr){2-6}
  & MASH-QA~\cite{zhu2020question}
  & Open-ended Q\&A
  & 34,808 consumer health Q\&A pairs from WebMD; extractive, multi-span answers (avg.\ 67 words) curated by healthcare experts from articles averaging 696 words
  & 19,989 & 1,000 \\
\cmidrule(lr){2-6}
  & ChatDoctor-iCliniq~\cite{li2023chatdoctor}
  & Open-ended Q\&A
  & 7,321 real patient--physician conversations from iCliniq.com spanning infectious disease, dermatology, cardiology, neurology, and other specialties
  & 6,321 & 1,000 \\
\midrule
\multirow{5}{*}{\parbox[c][25em][c]{\linewidth}{\centering Expert}}
  & BioASQ Task B~\cite{nentidis2025overview}
  & Summary Q\&A
  & Expert-curated biomedical questions paired with PubMed-grounded answers; covers yes/no, factoid, list, and summary types with gold concepts, snippets, and RDF triples
  & 1,283 & 80 \\
\cmidrule(lr){2-6}
  & MedQuAD~\cite{ben2019question}
  & Open-ended Q\&A
  & 47,457 Q\&A pairs from 12 NIH websites covering 37 question types (treatment, diagnosis, side effects, etc.) across diseases, drugs, and medical tests; enriched with UMLS CUI metadata
  & 15,407 & 1,000 \\
\cmidrule(lr){2-6}
  & MedQA-USMLE~\cite{jin2021disease}
  & Multiple-Choice Q\&A
  & \textasciitilde11,500 clinical vignette MCQs from the USMLE (Steps 1--3); each vignette describes patient demographics, symptoms, and history followed by a 4-option diagnostic or management question
  & 10,178 & 1,273 \\
\cmidrule(lr){2-6}
  & MedMCQA~\cite{pal2022medmcqa}
  & Multiple-Choice Q\&A
  & 194k+ MCQs from AIIMS \& NEET PG medical entrance exams spanning 21 subjects and 2,400+ topics (anatomy, pharmacology, pathology, surgery, psychiatry, etc.); includes answer explanations
  & 182,822 & 1,000 \\
\cmidrule(lr){2-6}
  & MMLU Medical~\cite{hendrycks2020measuring}
  & Multiple-Choice Q\&A
  & 1,242 four-choice questions from six MMLU medical subjects (clinical knowledge, medical genetics, anatomy, etc.), drawn from GRE and USMLE practice exams; evaluated in zero/few-shot settings
  & 642 & 600 \\
\bottomrule
\end{tabular}
\caption{Evaluation query datasets grouped by user type. \textit{Layman}: consumer health queries in everyday language; \textit{Expert}: biomedical/clinical questions. ``Query Set'': few-shot pool size; ``Test Set'': evaluation examples used.}
\label{tab:query-datasets}
\end{table*}

Datasets are grouped by intended user type: \textit{layman} datasets reflect consumer health inquiries in everyday language, while \textit{expert} datasets target biomedical professionals or medical students. The \textit{Query Set} column reports the eligible pool size used for few-shot prompting; the \textit{Test Set} column reports the number of examples used as evaluation queries. All corpora in Table~\ref{tab:retrieval-corpora} are indexed as whole records without further chunking; for Q\&A-format corpora (Yahoo Answers and HealthCareMagic) each document concatenates the question or title with the answer body.

\begin{table*}[t]
\centering
\scriptsize
\setlength{\tabcolsep}{3.5pt}
\renewcommand{\arraystretch}{1.15}

\begin{tabular}{C{1.5cm} C{3.2cm} C{5.8cm} C{1.5cm} C{1.7cm} C{1.2cm}}
\toprule
\textbf{User Type} & \textbf{Dataset} & \textbf{Source Content} & \textbf{Source Rows} & \textbf{Documents Retained} & \textbf{Dataset Link} \\
\midrule

\multirow{2}{*}{\parbox[c][5em][c]{\linewidth}{\centering Expert}}
& BioASQ Task A / PubMed~\cite{krithara2023bioasq}
& PubMed abstracts with human-assigned MeSH annotations (avg.\ 12.68 terms per article; up to 29,681 distinct MeSH terms across 16.2M articles)
& 16,218,838
& 16,218,838
& \datalink{https://participants-area.bioasq.org/Tasks/10a/trainingDataset/raw/allMeSH/} \\
\cmidrule(lr){2-6}
& Medical textbooks~\cite{xiong2024benchmarking}
& Retrieval-friendly chunks ($\leq$1,000 chars each) drawn from 18 authoritative biomedical textbooks spanning anatomy, physiology, pharmacology, pathology, and clinical medicine
& 125,847
& 125,847
& \datalink{https://huggingface.co/datasets/MedRAG/textbooks} \\
\midrule

\multirow{2}{*}{\parbox[c][5em][c]{\linewidth}{\centering Layman}}
& Yahoo Answers~\cite{yahoo_webscope_2009}
& Open-domain community Q\&A posts with best-answer labels and topic categories; user identities fully anonymized
& 1,400,000
& 1,238,506
& \datalink{https://www.kaggle.com/datasets/jarupula/yahoo-answers-dataset} \\
\cmidrule(lr){2-6}
& HealthCareMagic~\cite{li2023chatdoctor}
& Real-world patient symptom queries paired with detailed physician responses (diagnosis, treatment, and referral advice) across 10+ clinical specialties
& 112,165
& 112,165
& \datalink{https://huggingface.co/datasets/wangrongsheng/HealthCareMagic-100k-en} \\
\bottomrule
\end{tabular}

\caption{Retrieval corpora grouped by user type. \textit{Expert}: technical biomedical sources; \textit{Layman}: community health and general Q\&A. ``Documents Retained'': records after quality filtering.}
\label{tab:retrieval-corpora}
\end{table*}

\section{Prompt Templates}
\label{sec:appendix-prompts}

All generation experiments use two system prompts and two user message templates, combined via each model's native chat template.

\paragraph{Layman system prompt.} Used when the query dataset belongs to the \textit{layman} user type (MeQSum, MedRedQA, MedicationQA, MASH-QA, ChatDoctor-iCliniq):

\begin{tcolorbox}[
  colback=blue!5!white,
  colframe=blue!60!black,
  fontupper=\small,
  width=\linewidth,
  boxsep=4pt,
  title=\textbf{Layman System Prompt},
  coltitle=white,
  colbacktitle=blue!60!black
]
\textit{You are a helpful health assistant answering questions from members of the general public. Use simple, everyday language that a non-medical person can easily understand. Avoid medical jargon. Be clear, friendly, and concise.}
\end{tcolorbox}

\paragraph{Expert system prompt.} Used for \textit{expert} datasets (BioASQ Task~B, MedQuAD, MedQA-USMLE, MedMCQA, MMLU Medical):

\begin{tcolorbox}[
  colback=red!5!white,
  colframe=red!60!black,
  fontupper=\small,
  width=\linewidth,
  boxsep=4pt,
  title=\textbf{Expert System Prompt},
  coltitle=white,
  colbacktitle=red!60!black
]
\textit{You are a clinical decision support assistant. Answer questions from healthcare professionals using precise medical terminology. Provide evidence-based, clinically detailed responses with relevant diagnostic and therapeutic considerations.}
\end{tcolorbox}

\vspace{1mm} \noindent\textbf{User message (without retrieval)}

\begin{tcolorbox}[
  colback=green!5!white,
  colframe=green!50!black,
  fontupper=\small\ttfamily,
  width=\linewidth,
  boxsep=4pt,
  title=\textbf{User Message},
  coltitle=white,
  colbacktitle=green!50!black
]
QUESTION: \{query\}\\[4pt]
ANSWER:
\end{tcolorbox}

\vspace{1mm} \noindent\textbf{User message (with retrieval)}

\begin{tcolorbox}[
  colback=violet!5!white,
  colframe=violet!60!black,
  fontupper=\small\ttfamily,
  width=\linewidth,
  boxsep=4pt,
  title=\textbf{User Message},
  coltitle=white,
  colbacktitle=violet!60!black
]
Use the following retrieved passages to help answer the question.\\[4pt]
RETRIEVED CONTEXT:\\
\{context\}\\[4pt]
QUESTION: \{query\}\\[4pt]
ANSWER:
\end{tcolorbox}

\vspace{1mm} \noindent where \texttt{\{context\}} is a concatenation of the top-$k$ retrieved passages, each formatted as \texttt{[Passage N (source: \{source\})]: \{text\}}. The final prompt is produced by wrapping these system and user messages in each model's chat template.

\section{Full Results by Metric}
\label{sec:appendix-full-results}

This section provides per-dataset performance tables for all evaluated metrics beyond ROUGE-L (reported in the main paper). Tables~\ref{tab:results}--\ref{tab:results_bleu} report ROUGE-2, ROUGE-1, BERTScore, METEOR, and BLEU respectively, broken down by model and retrieval corpus across the seven open-ended datasets. Tables~\ref{tab:rouge1_retrieval}--\ref{tab:bert_retrieval} further break down ROUGE-1, ROUGE-2, BLEU, METEOR, and BERTScore by retrieval method (BM25, Hybrid, MedCPT, TF-IDF) across all four corpora.

\paragraph{ROUGE-2 (Table~\ref{tab:results}).}
The pattern mirrors ROUGE-L: the BioASQ/PubMed corpus produces the largest gains, and only on the BioASQ expert open-ended task. For example, LLaMA-3.1-8B improves from 14.03 to 18.34, and LLaMA-3.1-70B from 14.43 to 19.25, while all other datasets see gains under 1 point or negative effects. The average improvement over baseline is at most 0.53 points (LLaMA-3.1-8B: 5.82 $\to$ 6.35), and for Qwen2.5-72B Yahoo Answers produces the best average (6.39), marginally ahead of BioASQ (6.37), illustrating how small these corpus-level differences are.

\paragraph{ROUGE-1 (Table~\ref{tab:results_rouge1}).}
Again, the BioASQ corpus helps on the BioASQ dataset (gains of 3--5 points for all models) while effects on lay datasets are within $\pm$1 point. Mistral-7B shows an above-average improvement with BioASQ corpus on the BioASQ open-ended task (37.56 $\to$ 40.46 with BioASQ corpus; 40.53 with Qwen2.5-72B), confirming domain-matched retrieval has local benefit. Averaged over all datasets the maximum gain is 0.65 points (LLaMA-3.1-8B baseline 22.49 $\to$ best 22.88).

\paragraph{BERTScore (Table~\ref{tab:results_bert}).}
BERTScore is notably more stable than any ROUGE metric: the gap between the no-retrieval baseline and the best retrieval condition is under 0.7 points for all models. For instance, LLaMA-3.1-8B moves from 52.47 (baseline) to at best 52.85 (BioASQ corpus), a gain of just 0.38 points. This suggests that while retrieved context can slightly shift surface n-gram overlap, the overall semantic content of model outputs barely changes, consistent with the view that 7--8B models are not effectively incorporating the retrieved evidence.

\paragraph{METEOR (Table~\ref{tab:results_meteor}).}
METEOR shows small, mixed effects: the BioASQ corpus provides a modest boost on the BioASQ dataset (e.g., LLaMA-3.1-8B: 29.84 $\to$ 30.92; Qwen2.5-72B: 31.11 $\to$ 34.05), but on lay datasets retrieval often slightly lowers METEOR, particularly for the 70B models where the baseline exceeds all retrieval conditions on several tasks (e.g., LLaMA-3.1-70B average: 18.92 baseline vs.\ 17.03--17.12 across all corpora).

\paragraph{BLEU (Table~\ref{tab:results_bleu}).}
BLEU scores are generally very low for lay datasets ($< 2$ across all conditions), underscoring that n-gram precision is a weak signal for open-ended health QA. The BioASQ corpus produces notable gains on the expert BioASQ dataset (LLaMA-3.1-8B: 12.92 $\to$ 19.08; LLaMA-3.1-70B: 13.32 $\to$ 19.76), but all other datasets improve by less than 0.3 BLEU points, and many worsen. Average BLEU across all datasets improves by 0.82 points at most.

\paragraph{Retrieval method breakdown (Tables~\ref{tab:rouge1_retrieval}--\ref{tab:bert_retrieval}).}
Across all five metrics, differences among BM25, Hybrid (RRF), MedCPT, and TF-IDF are consistently within 0.5 metric points for any model--corpus combination. The Hybrid retriever shows a slight edge in several configurations (particularly ROUGE-L and ROUGE-1), while TF-IDF is competitive with BM25 despite its greater simplicity. No single retrieval method dominates across all metrics and models, reinforcing the conclusion that retrieval architecture choice is secondary to corpus and model selection.

\begin{table*}[t]
\centering
\resizebox{\textwidth}{!}{%
\begin{tabular}{llcccccccc}
\toprule
\textbf{Model} & \textbf{Retrieval Dataset} & \textbf{BioASQ} & \textbf{ChatDoctor/iCliniq} & \textbf{MashQA} & \textbf{MedicationQA} & \textbf{MedQuAD} & \textbf{MedRedQA} & \textbf{MeQSum} & \textbf{Average} \\
\midrule
\multirow{5}{*}{LLaMA3.1-8B}
 & w/o RAG            & 14.03 & 3.26 & 6.27 & 4.26 & 8.36 & 2.44 & 2.10 & 5.82 \\ \cmidrule(lr){3-10}
 & BioASQ             & 18.34 & 3.51 & 5.81 & 3.82 & 7.81 & 2.53 & 2.63 & 6.35 \\
 & HealthCareMagic    & 12.97 & 3.69 & 5.64 & 3.72 & 8.10 & 2.51 & 2.23 & 5.55 \\
 & Medical Textbooks  & 13.39 & 3.51 & 5.74 & 3.71 & 7.18 & 2.50 & 2.30 & 5.48 \\
 & Yahoo Answers      & 12.37 & 3.48 & 5.90 & 3.57 & 7.76 & 2.48 & 2.20 & 5.39 \\
\midrule
\multirow{5}{*}{LLaMA3.1-70B}
 & w/o RAG            & 14.43 & 3.55 & 7.41 & 5.42 & 9.32 & 2.40 & 1.88 & 6.34 \\\cmidrule(lr){3-10}
 & BioASQ             & 19.25 & 3.51 & 5.94 & 4.95 & 7.90 & 2.28 & 2.21 & 6.58 \\
 & HealthCareMagic    & 15.52 & 3.66 & 6.49 & 5.89 & 9.29 & 2.27 & 2.15 & 6.47 \\
 & Medical Textbooks  & 15.86 & 3.56 & 5.94 & 5.47 & 7.82 & 2.31 & 2.19 & 6.16 \\
 & Yahoo Answers      & 15.31 & 3.58 & 6.28 & 5.39 & 8.45 & 2.28 & 2.23 & 6.22 \\
\midrule
\multirow{5}{*}{Mistral-7B}
 & w/o RAG            & 14.18 & 3.54 & 6.48 & 4.03 & 7.96 & 2.53 & 2.09 & 5.83 \\\cmidrule(lr){3-10}
 & BioASQ             & 17.02 & 3.58 & 5.89 & 3.78 & 7.49 & 2.62 & 2.42 & 6.11 \\
 & HealthCareMagic    & 14.47 & 3.94 & 6.34 & 3.86 & 8.23 & 2.64 & 2.46 & 5.99 \\
 & Medical Textbooks  & 13.80 & 3.63 & 5.87 & 4.05 & 7.25 & 2.56 & 2.40 & 5.65 \\
 & Yahoo Answers      & 14.75 & 3.64 & 6.43 & 4.02 & 8.19 & 2.67 & 2.47 & 6.02 \\
\midrule
\multirow{5}{*}{Qwen2.5-7B}
 & w/o RAG            & 13.42 & 3.30 & 6.70 & 3.98 & 7.20 & 2.44 & 1.83 & 5.55 \\\cmidrule(lr){3-10}
 & BioASQ             & 15.68 & 3.38 & 6.50 & 3.77 & 7.19 & 2.48 & 2.13 & 5.88 \\
 & HealthCareMagic    & 13.32 & 3.37 & 6.27 & 3.50 & 7.43 & 2.49 & 1.96 & 5.48 \\
 & Medical Textbooks  & 14.02 & 3.44 & 6.40 & 3.87 & 6.95 & 2.45 & 2.14 & 5.61 \\
 & Yahoo Answers      & 13.48 & 3.37 & 6.47 & 3.79 & 7.45 & 2.49 & 2.19 & 5.61 \\
\midrule
\multirow{5}{*}{Qwen2.5-72B}
 & w/o RAG            & 14.60 & 3.49 & 7.10 & 3.94 & 8.64 & 2.79 & 1.86 & 6.06 \\\cmidrule(lr){3-10}
 & BioASQ             & 16.73 & 3.67 & 7.03 & 3.85 & 8.54 & 2.81 & 1.99 & 6.37 \\
 & HealthCareMagic    & 16.03 & 3.70 & 6.93 & 3.72 & 9.11 & 2.83 & 1.88 & 6.31 \\
 & Medical Textbooks  & 15.84 & 3.68 & 7.13 & 4.21 & 8.49 & 2.77 & 2.00 & 6.30 \\
 & Yahoo Answers      & 16.27 & 3.54 & 7.06 & 3.79 & 9.25 & 2.83 & 2.02 & 6.39 \\
\bottomrule
\end{tabular}%
}
\caption{ROUGE-2 by model and retrieval corpus (open-ended datasets).}
\label{tab:results}
\end{table*}

\begin{table*}[t]
\centering
\resizebox{\textwidth}{!}{%
\begin{tabular}{llcccccccc}
\toprule
\textbf{Model} & \textbf{Retrieval Dataset} & \textbf{BioASQ} & \textbf{ChatDoctor/iCliniq} & \textbf{MashQA} & \textbf{MedicationQA} & \textbf{MedQuAD} & \textbf{MedRedQA} & \textbf{MeQSum} & \textbf{Average} \\
\midrule
\multirow{5}{*}{LLaMA3.1-8B}
 & w/o RAG            & 36.34 & 22.91 & 24.67 & 20.15 & 30.51 & 15.46 & 7.36 & 22.49 \\\cmidrule(lr){3-10}
 & BioASQ             & 40.24 & 23.93 & 25.59 & 18.91 & 27.02 & 15.61 & 8.84 & 22.88 \\
 & HealthCareMagic    & 32.93 & 24.57 & 24.44 & 18.60 & 29.30 & 15.52 & 8.25 & 21.94 \\
 & Medical Textbooks  & 34.70 & 24.06 & 25.33 & 18.93 & 26.58 & 15.54 & 8.06 & 21.89 \\
 & Yahoo Answers      & 33.68 & 23.93 & 25.32 & 18.17 & 28.70 & 15.39 & 7.61 & 21.83 \\
\midrule
\multirow{5}{*}{LLaMA3.1-70B}
 & w/o RAG            & 36.80 & 24.87 & 28.54 & 22.23 & 31.57 & 15.73 & 6.76 & 23.79 \\\cmidrule(lr){3-10}
 & BioASQ             & 41.45 & 24.77 & 24.03 & 20.40 & 24.68 & 15.60 & 7.27 & 22.60 \\
 & HealthCareMagic    & 37.14 & 24.93 & 24.74 & 22.31 & 28.36 & 15.78 & 7.29 & 22.94 \\
 & Medical Textbooks  & 37.13 & 24.84 & 23.85 & 21.65 & 23.59 & 15.70 & 7.27 & 22.00 \\
 & Yahoo Answers      & 36.85 & 24.99 & 24.00 & 21.25 & 25.75 & 15.79 & 7.52 & 22.31 \\
\midrule
\multirow{5}{*}{Mistral-7B}
 & w/o RAG            & 37.56 & 25.83 & 27.02 & 20.79 & 31.15 & 16.38 & 7.70 & 23.78 \\\cmidrule(lr){3-10}
 & BioASQ             & 40.46 & 26.16 & 26.82 & 20.35 & 28.71 & 16.94 & 8.66 & 24.01 \\
 & HealthCareMagic    & 38.66 & 26.97 & 27.78 & 20.50 & 30.78 & 16.99 & 8.57 & 24.32 \\
 & Medical Textbooks  & 36.24 & 25.98 & 26.62 & 21.03 & 29.13 & 16.73 & 8.46 & 23.46 \\
 & Yahoo Answers      & 37.99 & 26.58 & 27.77 & 20.63 & 30.59 & 17.17 & 8.91 & 24.23 \\
\midrule
\multirow{5}{*}{Qwen2.5-7B}
 & w/o RAG            & 36.89 & 24.77 & 27.60 & 20.11 & 29.91 & 15.82 & 7.56 & 23.24 \\\cmidrule(lr){3-10}
 & BioASQ             & 38.95 & 24.79 & 27.49 & 20.09 & 29.51 & 15.76 & 7.71 & 23.47 \\
 & HealthCareMagic    & 36.58 & 24.67 & 27.22 & 19.21 & 30.42 & 15.78 & 7.60 & 23.07 \\
 & Medical Textbooks  & 37.09 & 24.78 & 27.41 & 20.29 & 29.33 & 15.86 & 7.85 & 23.23 \\
 & Yahoo Answers      & 36.88 & 24.88 & 27.95 & 20.00 & 30.47 & 15.87 & 7.84 & 23.41 \\
\midrule
\multirow{5}{*}{Qwen2.5-72B}
 & w/o RAG            & 38.75 & 25.07 & 28.29 & 19.96 & 31.91 & 16.35 & 7.64 & 24.00 \\\cmidrule(lr){3-10}
 & BioASQ             & 40.53 & 25.08 & 27.95 & 19.95 & 32.08 & 16.35 & 7.65 & 24.23 \\
 & HealthCareMagic    & 40.12 & 25.22 & 27.84 & 19.28 & 32.85 & 16.37 & 7.42 & 24.16 \\
 & Medical Textbooks  & 39.26 & 25.23 & 28.13 & 20.60 & 31.78 & 16.32 & 7.63 & 24.14 \\
 & Yahoo Answers      & 40.47 & 25.21 & 28.48 & 19.55 & 32.93 & 16.44 & 7.76 & 24.41 \\
\bottomrule
\end{tabular}%
}
\caption{ROUGE-1 by model and retrieval corpus (open-ended datasets).}
\label{tab:results_rouge1}
\end{table*}

\begin{table*}[t]
\centering
\resizebox{\textwidth}{!}{%
\begin{tabular}{llcccccccc}
\toprule
\textbf{Model} & \textbf{Retrieval Dataset} & \textbf{BioASQ} & \textbf{ChatDoctor/iCliniq} & \textbf{MashQA} & \textbf{MedicationQA} & \textbf{MedQuAD} & \textbf{MedRedQA} & \textbf{MeQSum} & \textbf{Average} \\
\midrule
\multirow{5}{*}{LLaMA3.1-8B}
 & w/o RAG            & 61.72 & 51.39 & 55.50 & 51.82 & 56.98 & 46.44 & 43.43 & 52.47 \\\cmidrule(lr){3-10}
 & BioASQ             & 62.88 & 52.44 & 55.04 & 50.84 & 56.88 & 47.07 & 44.78 & 52.85 \\
 & HealthCareMagic    & 57.43 & 52.56 & 54.25 & 49.98 & 56.72 & 47.17 & 43.52 & 51.66 \\
 & Medical Textbooks  & 59.87 & 52.40 & 54.81 & 50.53 & 56.03 & 47.05 & 44.45 & 52.16 \\
 & Yahoo Answers      & 58.37 & 52.29 & 55.13 & 49.41 & 56.61 & 47.03 & 43.98 & 51.83 \\
\midrule
\multirow{5}{*}{LLaMA3.1-70B}
 & w/o RAG            & 61.74 & 52.47 & 57.29 & 52.91 & 58.43 & 46.69 & 43.20 & 53.25 \\\cmidrule(lr){3-10}
 & BioASQ             & 63.96 & 52.40 & 54.45 & 52.27 & 56.84 & 46.63 & 44.30 & 52.97 \\
 & HealthCareMagic    & 60.41 & 52.58 & 55.53 & 53.46 & 58.35 & 46.55 & 44.36 & 53.03 \\
 & Medical Textbooks  & 61.08 & 52.44 & 54.78 & 53.15 & 56.45 & 46.53 & 44.34 & 52.68 \\
 & Yahoo Answers      & 60.87 & 52.31 & 55.02 & 52.54 & 57.38 & 46.57 & 44.37 & 52.72 \\
\midrule
\multirow{5}{*}{Mistral-7B}
 & w/o RAG            & 62.52 & 53.37 & 57.13 & 51.96 & 58.17 & 47.76 & 44.21 & 53.59 \\\cmidrule(lr){3-10}
 & BioASQ             & 63.06 & 53.49 & 55.84 & 51.23 & 57.33 & 48.08 & 45.52 & 53.51 \\
 & HealthCareMagic    & 61.77 & 54.04 & 56.87 & 51.41 & 58.29 & 48.10 & 45.20 & 53.67 \\
 & Medical Textbooks  & 61.05 & 53.43 & 56.09 & 52.10 & 57.10 & 48.05 & 45.20 & 53.29 \\
 & Yahoo Answers      & 61.76 & 53.47 & 56.77 & 51.22 & 58.11 & 48.20 & 45.32 & 53.55 \\
\midrule
\multirow{5}{*}{Qwen2.5-7B}
 & w/o RAG            & 61.28 & 52.02 & 55.81 & 50.94 & 56.07 & 46.07 & 42.91 & 52.16 \\\cmidrule(lr){3-10}
 & BioASQ             & 63.23 & 52.62 & 56.13 & 51.36 & 57.06 & 46.62 & 44.09 & 53.02 \\
 & HealthCareMagic    & 61.36 & 52.55 & 56.10 & 50.74 & 57.33 & 46.61 & 43.68 & 52.62 \\
 & Medical Textbooks  & 61.83 & 52.54 & 56.24 & 51.79 & 56.89 & 46.55 & 44.17 & 52.86 \\
 & Yahoo Answers      & 61.28 & 52.44 & 56.38 & 51.10 & 57.37 & 46.71 & 44.09 & 52.77 \\
\midrule
\multirow{5}{*}{Qwen2.5-72B}
 & w/o RAG            & 62.26 & 51.56 & 56.45 & 50.50 & 57.26 & 46.40 & 42.58 & 52.43 \\\cmidrule(lr){3-10}
 & BioASQ             & 64.30 & 51.92 & 56.60 & 50.98 & 57.86 & 46.63 & 43.34 & 53.09 \\
 & HealthCareMagic    & 63.05 & 52.07 & 56.40 & 50.26 & 58.04 & 46.69 & 42.94 & 52.78 \\
 & Medical Textbooks  & 63.07 & 51.93 & 56.81 & 51.54 & 57.75 & 46.57 & 43.29 & 52.99 \\
 & Yahoo Answers      & 63.56 & 51.88 & 56.73 & 50.53 & 58.20 & 46.72 & 43.33 & 52.99 \\
\bottomrule
\end{tabular}%
}
\caption{BERTScore by model and retrieval corpus (open-ended datasets).}
\label{tab:results_bert}
\end{table*}

\begin{table*}[t]
\centering
\resizebox{\textwidth}{!}{%
\begin{tabular}{llcccccccc}
\toprule
\textbf{Model} & \textbf{Retrieval Dataset} & \textbf{BioASQ} & \textbf{ChatDoctor/iCliniq} & \textbf{MashQA} & \textbf{MedicationQA} & \textbf{MedQuAD} & \textbf{MedRedQA} & \textbf{MeQSum} & \textbf{Average} \\
\midrule
\multirow{5}{*}{LLaMA3.1-8B}
 & w/o RAG            & 29.84 & 16.50 & 21.84 & 16.05 & 19.09 & 14.75 & 11.96 & 18.58 \\\cmidrule(lr){3-10}
 & BioASQ             & 30.92 & 18.70 & 18.82 & 13.82 & 15.97 & 15.26 & 13.02 & 18.07 \\
 & HealthCareMagic    & 25.69 & 19.11 & 18.89 & 13.64 & 18.37 & 15.32 & 12.18 & 17.60 \\
 & Medical Textbooks  & 26.58 & 18.73 & 18.44 & 12.92 & 15.40 & 15.18 & 12.69 & 17.13 \\
 & Yahoo Answers      & 25.73 & 18.92 & 19.28 & 13.36 & 17.66 & 15.21 & 12.22 & 17.48 \\
\midrule
\multirow{5}{*}{LLaMA3.1-70B}
 & w/o RAG            & 30.51 & 18.85 & 20.65 & 16.62 & 19.02 & 14.93 & 11.89 & 18.92 \\\cmidrule(lr){3-10}
 & BioASQ             & 31.90 & 18.37 & 15.08 & 13.76 & 13.04 & 14.37 & 12.67 & 17.03 \\
 & HealthCareMagic    & 28.27 & 18.32 & 15.92 & 14.63 & 15.86 & 14.25 & 12.59 & 17.12 \\
 & Medical Textbooks  & 27.75 & 18.45 & 14.77 & 13.60 & 12.16 & 14.42 & 12.48 & 16.23 \\
 & Yahoo Answers      & 28.76 & 18.35 & 14.40 & 13.01 & 14.09 & 14.22 & 12.68 & 16.50 \\
\midrule
\multirow{5}{*}{Mistral-7B}
 & w/o RAG            & 29.90 & 19.29 & 21.74 & 15.24 & 18.92 & 15.79 & 12.88 & 19.11 \\\cmidrule(lr){3-10}
 & BioASQ             & 31.67 & 19.31 & 18.94 & 14.06 & 16.13 & 15.79 & 14.19 & 18.58 \\
 & HealthCareMagic    & 29.29 & 19.45 & 19.85 & 13.98 & 18.09 & 15.91 & 13.94 & 18.64 \\
 & Medical Textbooks  & 28.42 & 19.40 & 19.41 & 14.61 & 16.69 & 15.78 & 14.00 & 18.33 \\
 & Yahoo Answers      & 30.11 & 19.25 & 19.49 & 13.85 & 17.76 & 15.82 & 14.49 & 18.68 \\
\midrule
\multirow{5}{*}{Qwen2.5-7B}
 & w/o RAG            & 30.05 & 19.22 & 22.34 & 14.73 & 18.89 & 15.69 & 12.69 & 19.09 \\\cmidrule(lr){3-10}
 & BioASQ             & 31.30 & 19.35 & 20.66 & 14.56 & 18.15 & 15.74 & 13.05 & 18.97 \\
 & HealthCareMagic    & 29.74 & 19.28 & 20.91 & 14.10 & 19.02 & 15.68 & 12.91 & 18.81 \\
 & Medical Textbooks  & 30.02 & 19.40 & 20.46 & 14.58 & 17.75 & 15.77 & 13.28 & 18.75 \\
 & Yahoo Answers      & 29.19 & 19.09 & 20.49 & 14.09 & 19.21 & 15.63 & 13.38 & 18.73 \\
\midrule
\multirow{5}{*}{Qwen2.5-72B}
 & w/o RAG            & 31.11 & 19.12 & 22.27 & 14.45 & 20.09 & 15.98 & 12.60 & 19.37 \\\cmidrule(lr){3-10}
 & BioASQ             & 34.05 & 19.44 & 22.01 & 15.07 & 19.86 & 16.05 & 12.97 & 19.92 \\
 & HealthCareMagic    & 33.11 & 19.58 & 22.18 & 14.69 & 20.62 & 16.07 & 12.60 & 19.84 \\
 & Medical Textbooks  & 32.75 & 19.45 & 22.15 & 15.30 & 19.86 & 16.04 & 12.82 & 19.77 \\
 & Yahoo Answers      & 33.44 & 19.24 & 21.90 & 14.36 & 20.73 & 16.10 & 12.97 & 19.82 \\
\bottomrule
\end{tabular}%
}
\caption{METEOR by model and retrieval corpus (open-ended datasets).}
\label{tab:results_meteor}
\end{table*}

\begin{table*}[t]
\centering
\resizebox{\textwidth}{!}{%
\begin{tabular}{llcccccccc}
\toprule
\textbf{Model} & \textbf{Retrieval Dataset} & \textbf{BioASQ} & \textbf{ChatDoctor/iCliniq} & \textbf{MashQA} & \textbf{MedicationQA} & \textbf{MedQuAD} & \textbf{MedRedQA} & \textbf{MeQSum} & \textbf{Average} \\
\midrule
\multirow{5}{*}{LLaMA3.1-8B}
 & w/o RAG            & 12.92 & 0.72 & 1.42 & 1.13 & 2.54 & 0.47 & 0.40 & 2.80 \\\cmidrule(lr){3-10}
 & BioASQ             & 19.08 & 0.77 & 1.52 & 1.00 & 1.99 & 0.48 & 0.53 & 3.62 \\
 & HealthCareMagic    & 10.52 & 0.82 & 1.39 & 0.99 & 2.34 & 0.48 & 0.50 & 2.43 \\
 & Medical Textbooks  & 12.99 & 0.78 & 1.49 & 1.02 & 1.83 & 0.48 & 0.49 & 2.73 \\
 & Yahoo Answers      & 10.79 & 0.76 & 1.53 & 0.97 & 2.16 & 0.47 & 0.44 & 2.45 \\
\midrule
\multirow{5}{*}{LLaMA3.1-70B}
 & w/o RAG            & 13.32 & 0.84 & 2.03 & 1.42 & 3.08 & 0.47 & 0.31 & 3.07 \\\cmidrule(lr){3-10}
 & BioASQ             & 19.76 & 0.85 & 1.56 & 1.19 & 1.73 & 0.49 & 0.39 & 3.71 \\
 & HealthCareMagic    & 14.16 & 0.92 & 1.72 & 1.57 & 2.58 & 0.48 & 0.44 & 3.12 \\
 & Medical Textbooks  & 15.92 & 0.87 & 1.47 & 1.53 & 1.67 & 0.48 & 0.41 & 3.19 \\
 & Yahoo Answers      & 13.97 & 0.92 & 1.54 & 1.43 & 2.03 & 0.49 & 0.47 & 2.98 \\
\midrule
\multirow{5}{*}{Mistral-7B}
 & w/o RAG            & 13.09 & 0.76 & 1.62 & 1.06 & 2.28 & 0.47 & 0.35 & 2.80 \\\cmidrule(lr){3-10}
 & BioASQ             & 17.25 & 0.83 & 1.67 & 1.07 & 2.12 & 0.51 & 0.45 & 3.41 \\
 & HealthCareMagic    & 13.74 & 0.96 & 1.69 & 1.12 & 2.35 & 0.53 & 0.43 & 2.97 \\
 & Medical Textbooks  & 12.71 & 0.81 & 1.58 & 1.15 & 2.01 & 0.50 & 0.45 & 2.74 \\
 & Yahoo Answers      & 14.32 & 0.86 & 1.75 & 1.14 & 2.44 & 0.52 & 0.46 & 3.07 \\
\midrule
\multirow{5}{*}{Qwen2.5-7B}
 & w/o RAG            & 11.83 & 0.68 & 1.65 & 1.03 & 1.99 & 0.43 & 0.33 & 2.56 \\\cmidrule(lr){3-10}
 & BioASQ             & 15.71 & 0.73 & 1.77 & 1.04 & 2.11 & 0.45 & 0.37 & 3.17 \\
 & HealthCareMagic    & 11.85 & 0.71 & 1.61 & 0.94 & 2.13 & 0.45 & 0.35 & 2.58 \\
 & Medical Textbooks  & 13.12 & 0.74 & 1.68 & 1.07 & 1.97 & 0.45 & 0.39 & 2.77 \\
 & Yahoo Answers      & 11.51 & 0.73 & 1.66 & 1.06 & 2.15 & 0.47 & 0.38 & 2.57 \\
\midrule
\multirow{5}{*}{Qwen2.5-72B}
 & w/o RAG            & 13.58 & 0.72 & 1.84 & 1.12 & 2.65 & 0.49 & 0.31 & 2.96 \\\cmidrule(lr){3-10}
 & BioASQ             & 16.43 & 0.76 & 1.83 & 1.10 & 2.59 & 0.51 & 0.34 & 3.37 \\
 & HealthCareMagic    & 14.51 & 0.78 & 1.80 & 0.99 & 2.88 & 0.52 & 0.32 & 3.11 \\
 & Medical Textbooks  & 14.67 & 0.77 & 1.89 & 1.12 & 2.62 & 0.50 & 0.34 & 3.13 \\
 & Yahoo Answers      & 14.79 & 0.75 & 1.92 & 1.04 & 2.95 & 0.53 & 0.36 & 3.19 \\
\bottomrule
\end{tabular}%
}
\caption{BLEU by model and retrieval corpus (open-ended datasets).}
\label{tab:results_bleu}
\end{table*}

\begin{table}[t]
\centering
\resizebox{\columnwidth}{!}{%
\begin{tabular}{llccccc}
\toprule
\textbf{Model} & \textbf{Retrieval Dataset} & \textbf{BM25} & \textbf{Hybrid} & \textbf{MedCPT} & \textbf{TF-IDF} \\
\midrule
\multirow{5}{*}{LLaMA3.1-8B}
 & BioASQ            & 22.20 & 22.63 & 22.32 & 21.75 \\
 & HealthCareMagic   & 21.09 & 21.70 & 21.48 & 21.48 \\
 & Medical Textbooks & 21.48 & 21.31 & 21.75 & 21.40 \\
 & Yahoo Answers     & 21.30 & 21.66 & 21.43 & 20.55 \\
\cmidrule(lr){2-6}
 & Average           & 21.52 & 21.83 & 21.74 & 21.30 \\
\midrule
\multirow{5}{*}{LLaMA3.1-70B}
 & BioASQ            & 21.58 & 22.40 & 21.82 & 21.68 \\
 & HealthCareMagic   & 22.50 & 22.63 & 22.90 & 22.62 \\
 & Medical Textbooks & 21.30 & 21.54 & 21.81 & 21.56 \\
 & Yahoo Answers     & 21.64 & 21.98 & 22.25 & 21.56 \\
\cmidrule(lr){2-6}
 & Average           & 21.76 & 22.14 & 22.20 & 21.85 \\
\midrule
\multirow{5}{*}{Mistral-7B}
 & BioASQ            & 23.60 & 23.88 & 23.59 & 23.31 \\
 & HealthCareMagic   & 23.98 & 23.97 & 24.00 & 24.07 \\
 & Medical Textbooks & 23.31 & 23.35 & 23.20 & 23.32 \\
 & Yahoo Answers     & 23.88 & 23.90 & 24.06 & 24.01 \\
\cmidrule(lr){2-6}
 & Average           & 23.69 & 23.77 & 23.71 & 23.68 \\
\midrule
\multirow{5}{*}{Qwen2.5-7B}
 & BioASQ            & 23.05 & 23.34 & 23.18 & 22.98 \\
 & HealthCareMagic   & 22.84 & 22.97 & 22.91 & 22.79 \\
 & Medical Textbooks & 22.87 & 23.18 & 23.01 & 23.14 \\
 & Yahoo Answers     & 23.00 & 23.17 & 23.39 & 23.24 \\
\cmidrule(lr){2-6}
 & Average           & 22.94 & 23.17 & 23.12 & 23.04 \\
 \midrule
 \multirow{5}{*}{Qwen2.5-72B}
 & BioASQ            & 24.03 & 24.09 & 23.91 & 24.01 \\
 & HealthCareMagic   & 23.83 & 23.91 & 23.90 & 24.13 \\
 & Medical Textbooks & 24.03 & 24.05 & 23.90 & 23.97 \\
 & Yahoo Answers     & 24.00 & 24.11 & 24.37 & 24.22 \\
\cmidrule(lr){2-6}
 & Average           & 23.97 & 24.04 & 24.02 & 24.08 \\
\bottomrule
\end{tabular}%
}
\caption{ROUGE-1 by retrieval method.}
\label{tab:rouge1_retrieval}
\end{table}

\begin{table}[t]
\centering
\resizebox{\columnwidth}{!}{%
\begin{tabular}{llccccc}
\toprule
\textbf{Model} & \textbf{Retrieval Dataset} & \textbf{BM25} & \textbf{Hybrid} & \textbf{MedCPT} & \textbf{TF-IDF} \\
\midrule
\multirow{5}{*}{LLaMA3.1-8B}
 & BioASQ            & 6.03 & 6.25 & 5.74 & 5.51 \\
 & HealthCareMagic   & 5.16 & 5.41 & 5.36 & 5.37 \\
 & Medical Textbooks & 5.27 & 5.33 & 5.41 & 5.33 \\
 & Yahoo Answers     & 5.27 & 5.34 & 5.26 & 5.12 \\
\cmidrule(lr){2-6}
 & Average           & 5.43 & 5.58 & 5.44 & 5.33 \\
\midrule
\multirow{5}{*}{LLaMA3.1-70B}
 & BioASQ            & 6.00 & 6.50 & 5.87 & 5.73 \\
 & HealthCareMagic   & 6.11 & 6.31 & 6.46 & 6.19 \\
 & Medical Textbooks & 5.70 & 5.81 & 6.01 & 5.97 \\
 & Yahoo Answers     & 5.90 & 6.00 & 6.03 & 5.71 \\
\cmidrule(lr){2-6}
 & Average           & 5.93 & 6.16 & 6.10 & 5.90 \\
\midrule
\multirow{5}{*}{Mistral-7B}
 & BioASQ            & 5.85 & 6.03 & 5.62 & 5.64 \\
 & HealthCareMagic   & 5.81 & 5.86 & 5.86 & 5.91 \\
 & Medical Textbooks & 5.52 & 5.61 & 5.50 & 5.50 \\
 & Yahoo Answers     & 5.91 & 5.93 & 5.91 & 5.91 \\
\cmidrule(lr){2-6}
 & Average           & 5.77 & 5.86 & 5.72 & 5.74 \\
\midrule
\multirow{5}{*}{Qwen2.5-7B}
 & BioASQ            & 5.48 & 5.77 & 5.52 & 5.36 \\
 & HealthCareMagic   & 5.25 & 5.36 & 5.35 & 5.36 \\
 & Medical Textbooks & 5.34 & 5.59 & 5.45 & 5.45 \\
 & Yahoo Answers     & 5.47 & 5.52 & 5.47 & 5.48 \\
\cmidrule(lr){2-6}
 & Average           & 5.39 & 5.56 & 5.45 & 5.41 \\
 \midrule
 \multirow{5}{*}{Qwen2.5-72B}
 & BioASQ            & 6.22 & 6.27 & 6.03 & 6.21 \\
 & HealthCareMagic   & 6.10 & 6.18 & 6.24 & 6.26 \\
 & Medical Textbooks & 6.24 & 6.21 & 6.05 & 6.22 \\
 & Yahoo Answers     & 6.18 & 6.22 & 6.33 & 6.35 \\
\cmidrule(lr){2-6}
 & Average           & 6.19 & 6.22 & 6.16 & 6.26 \\
\bottomrule
\end{tabular}%
}
\caption{ROUGE-2 by retrieval method.}
\label{tab:rouge2_retrieval}
\end{table}

\begin{table}[t]
\centering
\resizebox{\columnwidth}{!}{%
\begin{tabular}{llccccc}
\toprule
\textbf{Model} & \textbf{Retrieval Dataset} & \textbf{BM25} & \textbf{Hybrid} & \textbf{MedCPT} & \textbf{TF-IDF} \\
\midrule
\multirow{5}{*}{LLaMA3.1-8B}
 & BioASQ            & 3.32 & 3.59 & 3.10 & 2.81 \\
 & HealthCareMagic   & 2.11 & 2.36 & 2.38 & 2.31 \\
 & Medical Textbooks & 2.47 & 2.61 & 2.70 & 2.48 \\
 & Yahoo Answers     & 2.42 & 2.43 & 2.34 & 2.20 \\
\cmidrule(lr){2-6}
 & Average           & 2.58 & 2.75 & 2.63 & 2.45 \\
\midrule
\multirow{5}{*}{LLaMA3.1-70B}
 & BioASQ            & 3.26 & 3.67 & 3.15 & 2.94 \\
 & HealthCareMagic   & 2.90 & 2.89 & 3.12 & 2.94 \\
 & Medical Textbooks & 2.81 & 2.80 & 2.85 & 3.10 \\
 & Yahoo Answers     & 2.73 & 2.86 & 2.96 & 2.77 \\
\cmidrule(lr){2-6}
 & Average           & 2.93 & 3.06 & 3.02 & 2.94 \\
\midrule
\multirow{5}{*}{Mistral-7B}
 & BioASQ            & 3.19 & 3.38 & 2.93 & 2.68 \\
 & HealthCareMagic   & 2.77 & 2.89 & 2.92 & 2.94 \\
 & Medical Textbooks & 2.66 & 2.70 & 2.69 & 2.56 \\
 & Yahoo Answers     & 2.95 & 3.01 & 2.95 & 2.98 \\
\cmidrule(lr){2-6}
 & Average           & 2.89 & 3.00 & 2.87 & 2.79 \\
\midrule
\multirow{5}{*}{Qwen2.5-7B}
 & BioASQ            & 2.76 & 3.14 & 2.67 & 2.46 \\
 & HealthCareMagic   & 2.39 & 2.45 & 2.54 & 2.50 \\
 & Medical Textbooks & 2.53 & 2.76 & 2.56 & 2.46 \\
 & Yahoo Answers     & 2.52 & 2.54 & 2.51 & 2.55 \\
\cmidrule(lr){2-6}
 & Average           & 2.55 & 2.72 & 2.57 & 2.49 \\
 \midrule
 \multirow{5}{*}{Qwen2.5-72B}
 & BioASQ            & 3.24 & 3.34 & 3.05 & 3.06 \\
 & HealthCareMagic   & 2.95 & 2.95 & 3.10 & 3.10 \\
 & Medical Textbooks & 3.12 & 3.01 & 2.93 & 3.01 \\
 & Yahoo Answers     & 3.01 & 3.11 & 3.14 & 3.18 \\
\cmidrule(lr){2-6}
 & Average           & 3.08 & 3.10 & 3.06 & 3.09 \\
\bottomrule
\end{tabular}%
}
\caption{BLEU by retrieval method.}
\label{tab:bleu_retrieval}
\end{table}

\begin{table}[t]
\centering
\resizebox{\columnwidth}{!}{%
\begin{tabular}{llccccc}
\toprule
\textbf{Model} & \textbf{Retrieval Dataset} & \textbf{BM25} & \textbf{Hybrid} & \textbf{MedCPT} & \textbf{TF-IDF} \\
\midrule
\multirow{5}{*}{LLaMA3.1-8B}
 & BioASQ            & 17.19 & 17.54 & 17.55 & 17.27 \\
 & HealthCareMagic   & 16.53 & 17.02 & 17.17 & 17.05 \\
 & Medical Textbooks & 16.77 & 16.76 & 17.06 & 16.96 \\
 & Yahoo Answers     & 17.04 & 17.32 & 17.27 & 16.62 \\
\cmidrule(lr){2-6}
 & Average           & 16.88 & 17.16 & 17.26 & 16.98 \\
\midrule
\multirow{5}{*}{LLaMA3.1-70B}
 & BioASQ            & 16.11 & 16.55 & 16.32 & 16.17 \\
 & HealthCareMagic   & 16.56 & 16.70 & 17.07 & 16.82 \\
 & Medical Textbooks & 15.65 & 15.77 & 16.17 & 15.86 \\
 & Yahoo Answers     & 15.94 & 16.04 & 16.36 & 16.08 \\
\cmidrule(lr){2-6}
 & Average           & 16.06 & 16.26 & 16.48 & 16.23 \\
\midrule
\multirow{5}{*}{Mistral-7B}
 & BioASQ            & 18.03 & 18.38 & 18.13 & 18.00 \\
 & HealthCareMagic   & 18.33 & 18.36 & 18.44 & 18.45 \\
 & Medical Textbooks & 18.12 & 18.08 & 18.14 & 18.12 \\
 & Yahoo Answers     & 18.34 & 18.39 & 18.31 & 18.55 \\
\cmidrule(lr){2-6}
 & Average           & 18.20 & 18.30 & 18.25 & 18.28 \\
\midrule
\multirow{5}{*}{Qwen2.5-7B}
 & BioASQ            & 18.37 & 18.74 & 18.59 & 18.43 \\
 & HealthCareMagic   & 18.48 & 18.44 & 18.70 & 18.61 \\
 & Medical Textbooks & 18.40 & 18.67 & 18.49 & 18.44 \\
 & Yahoo Answers     & 18.53 & 18.46 & 18.64 & 18.59 \\
\cmidrule(lr){2-6}
 & Average           & 18.44 & 18.58 & 18.61 & 18.52 \\
 \midrule
 \multirow{5}{*}{Qwen2.5-72B}
 & BioASQ            & 19.63 & 19.76 & 19.56 & 19.52 \\
 & HealthCareMagic   & 19.55 & 19.65 & 19.81 & 19.57 \\
 & Medical Textbooks & 19.65 & 19.61 & 19.44 & 19.61 \\
 & Yahoo Answers     & 19.52 & 19.53 & 19.65 & 19.71 \\
\cmidrule(lr){2-6}
 & Average           & 19.59 & 19.64 & 19.61 & 19.60 \\
\bottomrule
\end{tabular}%
}
\caption{METEOR by retrieval method.}
\label{tab:meteor_retrieval}
\end{table}

\begin{table}[t]
\centering
\resizebox{\columnwidth}{!}{%
\begin{tabular}{llccccc}
\toprule
\textbf{Model} & \textbf{Retrieval Dataset} & \textbf{BM25} & \textbf{Hybrid} & \textbf{MedCPT} & \textbf{TF-IDF} \\
\midrule
\multirow{5}{*}{LLaMA3.1-8B}
 & BioASQ            & 52.48 & 52.72 & 52.54 & 52.08 \\
 & HealthCareMagic   & 50.92 & 51.33 & 51.24 & 51.37 \\
 & Medical Textbooks & 51.75 & 51.82 & 52.14 & 51.73 \\
 & Yahoo Answers     & 51.43 & 51.71 & 51.53 & 50.80 \\
\cmidrule(lr){2-6}
 & Average           & 51.65 & 51.89 & 51.86 & 51.49 \\
\midrule
\multirow{5}{*}{LLaMA3.1-70B}
 & BioASQ            & 52.36 & 52.88 & 52.52 & 52.07 \\
 & HealthCareMagic   & 52.69 & 52.84 & 52.94 & 52.78 \\
 & Medical Textbooks & 52.02 & 52.38 & 52.62 & 52.25 \\
 & Yahoo Answers     & 52.25 & 52.62 & 52.65 & 52.03 \\
\cmidrule(lr){2-6}
 & Average           & 52.33 & 52.68 & 52.68 & 52.28 \\
\midrule
\multirow{5}{*}{Mistral-7B}
 & BioASQ            & 53.25 & 53.39 & 53.24 & 53.04 \\
 & HealthCareMagic   & 53.47 & 53.55 & 53.53 & 53.56 \\
 & Medical Textbooks & 53.12 & 53.16 & 53.04 & 53.03 \\
 & Yahoo Answers     & 53.40 & 53.43 & 53.50 & 53.39 \\
\cmidrule(lr){2-6}
 & Average           & 53.31 & 53.39 & 53.33 & 53.26 \\
\midrule
\multirow{5}{*}{Qwen2.5-7B}
 & BioASQ            & 52.72 & 52.96 & 52.80 & 52.53 \\
 & HealthCareMagic   & 52.37 & 52.45 & 52.51 & 52.37 \\
 & Medical Textbooks & 52.52 & 52.81 & 52.73 & 52.64 \\
 & Yahoo Answers     & 52.65 & 52.64 & 52.75 & 52.60 \\
\cmidrule(lr){2-6}
 & Average           & 52.57 & 52.71 & 52.70 & 52.54 \\
 \midrule
 \multirow{5}{*}{Qwen2.5-72B}
 & BioASQ            & 52.93 & 53.01 & 52.87 & 52.90 \\
 & HealthCareMagic   & 52.60 & 52.70 & 52.74 & 52.69 \\
 & Medical Textbooks & 52.89 & 52.87 & 52.84 & 52.85 \\
 & Yahoo Answers     & 52.74 & 52.85 & 52.92 & 52.86 \\
\cmidrule(lr){2-6}
 & Average           & 52.79 & 52.86 & 52.84 & 52.83 \\
\bottomrule
\end{tabular}%
}
\caption{BERTScore by retrieval method.}
\label{tab:bert_retrieval}
\end{table}

\section{Ablation Study: Additional Figures}
\label{sec:appendix-ablations}

\begin{figure*}[t]
    \centering
    \includegraphics[width=\linewidth]{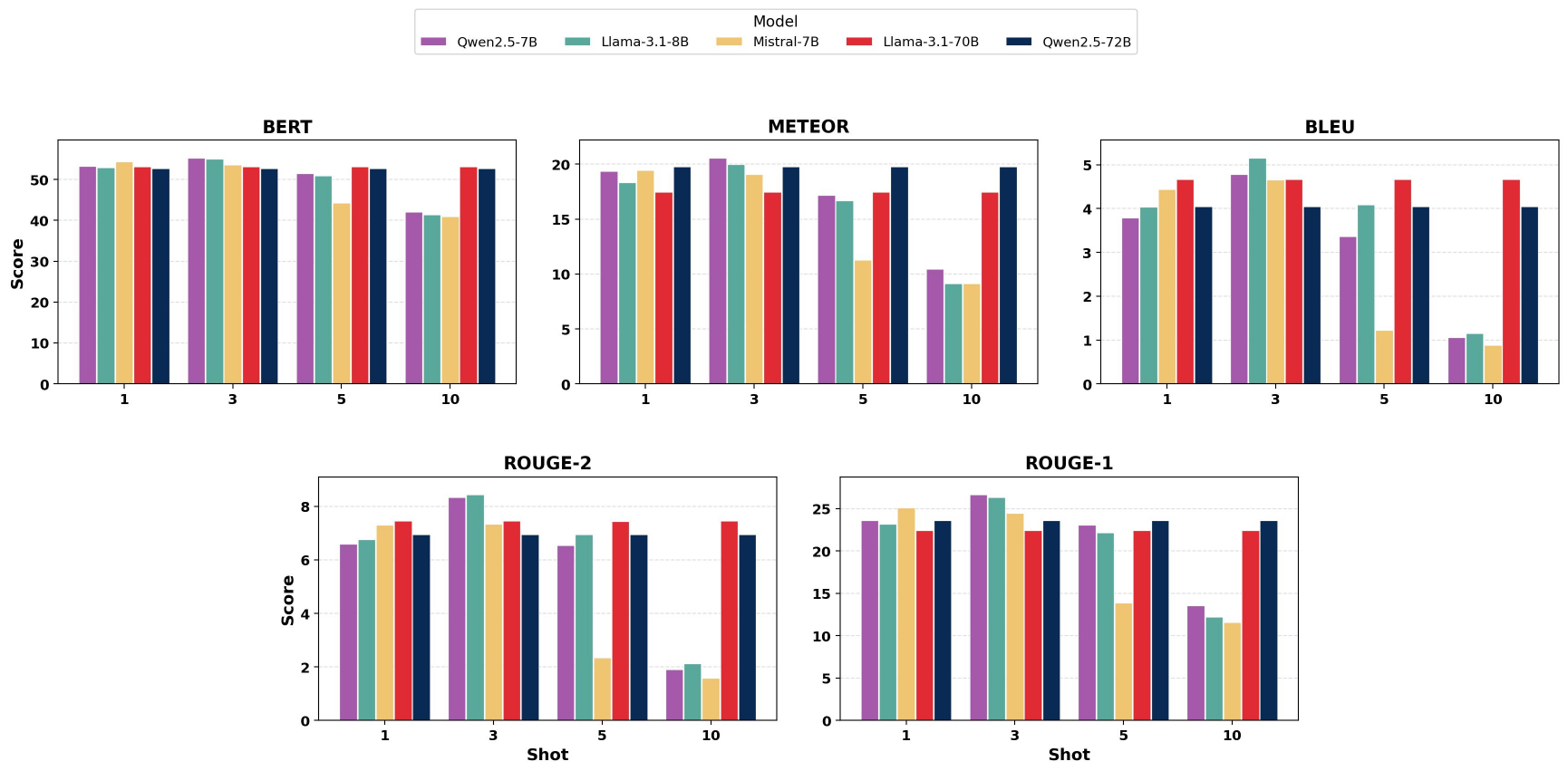}
    \caption{Additional open-ended metrics across shot counts (few-shot ablation).}
    \label{fig:open-few-shot-appeindex}
\end{figure*}

\begin{figure*}[t]
    \centering
    \includegraphics[width=\linewidth]{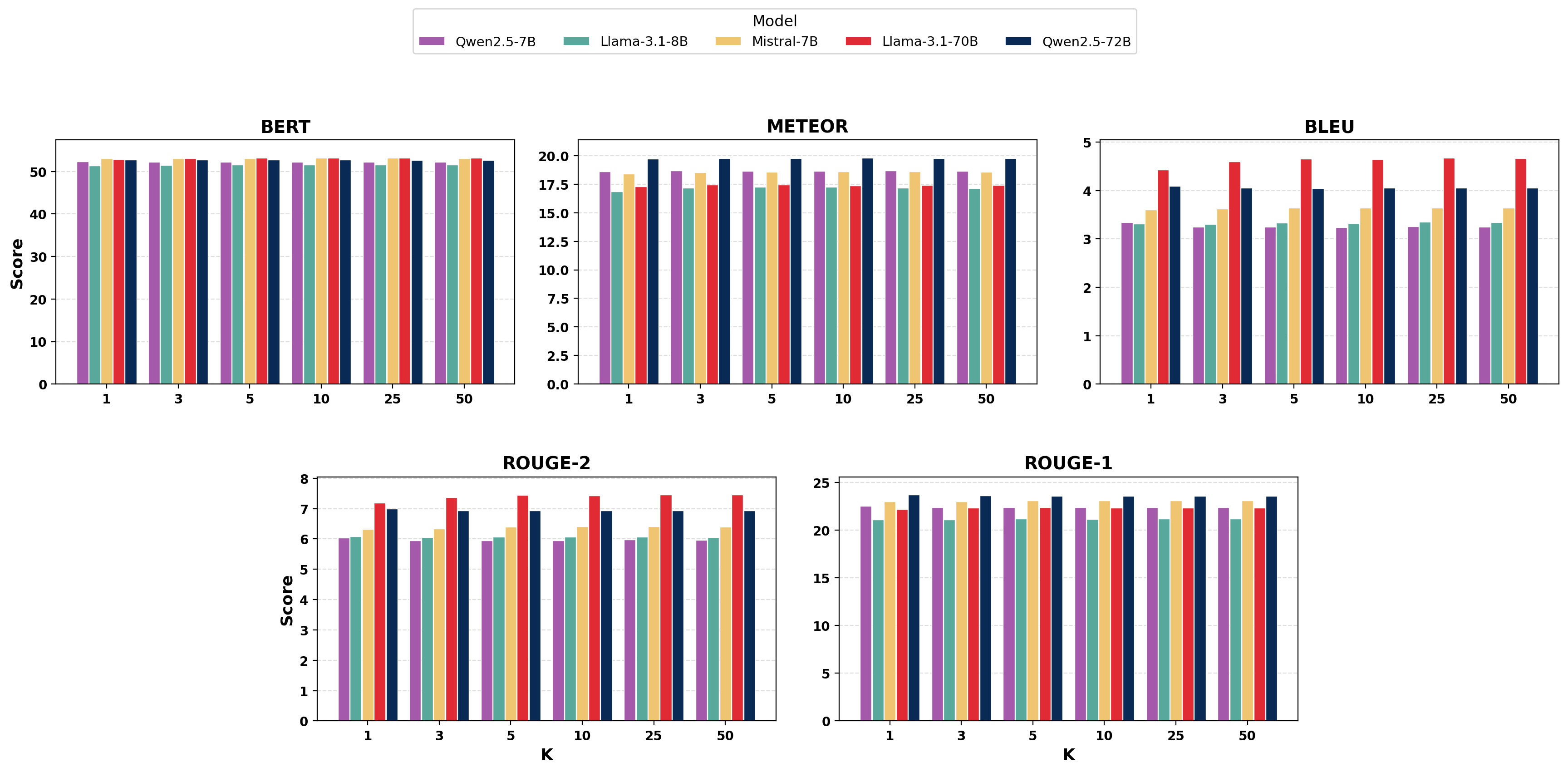}
    \caption{Additional open-ended metrics across top-$k$ values.}
    \label{fig:open-k-value-appeindex}
\end{figure*}

This section provides additional figures for the two ablation studies described in Section~\ref{sec:ablation}. Figure~\ref{fig:open-few-shot-appeindex} shows BERTScore, METEOR, BLEU, ROUGE-2, and ROUGE-1 trends under few-shot prompting across all five models on open-ended questions. Figure~\ref{fig:open-k-value-appeindex} shows the same metrics across top-$k$ values.

\paragraph{Few-shot: additional metrics (Figure~\ref{fig:open-few-shot-appeindex}).}
All six open-ended metrics tell a consistent story. For the larger models (LLaMA-3.1-70B and Qwen2.5-72B), all metrics are flat across all shot counts: for example, METEOR stays at 17.44--17.45 for LLaMA-3.1-70B and BERTScore stays at 52.69 for Qwen2.5-72B regardless of shot count. For smaller models, the 3-shot sweet spot and subsequent collapse are visible in every metric. Specifically, ROUGE-1 peaks at 3 shots for LLaMA-3.1-8B (26.29) and Qwen2.5-7B (26.61) before collapsing to 12.19 and 13.52 at 10 shots. METEOR follows the same pattern: LLaMA-3.1-8B peaks at 19.99 (3 shots) vs.\ 9.12 (10 shots), and Qwen2.5-7B at 20.54 (3 shots) vs.\ 10.43 (10 shots). BERTScore shows a more severe drop for Mistral-7B: from 54.32 at 1 shot to 40.96 at 10 shots, a 13-point collapse. BLEU, while numerically small, also collapses dramatically, LLaMA-3.1-8B drops from 5.15 (3 shots) to 1.15 (10 shots), and Mistral-7B from 4.66 (3 shots) to 0.88 (10 shots), confirming that higher shot counts severely degrade output quality at this scale.

\paragraph{Top-$k$: additional metrics (Figure~\ref{fig:open-k-value-appeindex}).}
The plateau behavior seen in ROUGE-L (main paper) extends to all six additional metrics. ROUGE-1 changes by at most 0.13 points from $k{=}5$ to $k{=}50$ across all models. ROUGE-2 is similarly stable: for example, LLaMA-3.1-70B moves from 7.44 ($k{=}5$) to 7.45 ($k{=}50$), a negligible change. METEOR plateaus by $k{=}5$ for most models, with variations under 0.1 between $k{=}5$ and $k{=}50$. BERTScore is the most stable metric of all: for LLaMA-3.1-8B, it ranges only from 51.33 ($k{=}1$) to 51.53 ($k{=}10$), a 0.20-point spread across all six $k$ values. The transition from $k{=}1$ to $k{=}5$ accounts for nearly all the variation, and additional passages beyond $k{=}5$ provide no measurable benefit in any metric for any model.

\end{document}